%% file: arxiv_main.tex
\PassOptionsToPackage{table}{xcolor}
\documentclass[runningheads]{llncs}

\usepackage{eccv}

\usepackage{eccvabbrv}

\usepackage{graphicx}
\usepackage{subcaption}
\usepackage{booktabs}
\usepackage{amsmath}
\usepackage{amssymb}
\usepackage{mathtools}
\usepackage{pifont}
\usepackage{enumitem}
\usepackage[most]{tcolorbox}
\usepackage{array}
\usepackage{arydshln}
\usepackage{tabularx}
\usepackage{multirow}
\usepackage{float}

\definecolor{takeawayblue}{RGB}{230,243,255}
\definecolor{titleblue}{RGB}{78,146,222}
\definecolor{lightblue}{RGB}{232,240,255}
\definecolor{deepblue}{RGB}{160,190,255}
\definecolor{lightgrey}{RGB}{234,234,234}
\definecolor{lightyellow}{RGB}{255,251,216}
\definecolor{mygreen}{RGB}{31,229,113}

\newcommand{\icohalf}{\textcolor{orange}{\ding{51}\kern-0.50em\ding{55}}}

\newtcolorbox{takeawaybox}[1][]{
  colback=takeawayblue!60,
  colframe=titleblue,
  coltitle=white,
  fonttitle=\footnotesize,
  enhanced,
  boxrule=0pt,
  frame hidden,
  sharp corners,
  breakable,
  title={\textcolor{white}{Key Takeaways: #1}},
  attach boxed title to top left={yshift=-1.15mm,xshift=0mm},
  boxed title style={
    colback=titleblue,
    sharp corners,
    boxrule=0pt,
    interior style={fill=titleblue,opacity=1},
    breakable,
    text width=\linewidth-1.5cm,
  },
  before skip=8pt,
  after skip=8pt,
  left=0mm, right=2mm, top=1mm, bottom=0mm,
}

\newtcolorbox{promptbox}[2][]{
    colback=white,
    coltext=black,
    arc=3mm,
    boxrule=0.5pt,
    colframe=black!60!white,
    title={#2},
    colbacktitle=black,
    coltitle=white,
    fonttitle=\bfseries,
    top=8pt,
    bottom=8pt,
    left=10pt,
    right=10pt,
    breakable,
    before upper={%
        \linespread{1}\selectfont
        \setlength{\parskip}{1ex plus 0.2ex minus 0.2ex}%
        \setlength{\parindent}{0pt}%
    },
    #1
}

\usepackage[hidelinks]{hyperref}

\usepackage{orcidlink}

\usepackage[capitalize,noabbrev]{cleveref}

\begin{document}

% ---------------------------------------------------------------
\ifdefined\SUPPLEMENTONLY
\title{Supplementary Material for ``From Evaluation to Enhancement:
Benchmarking and Improving Think-with-Video Reasoning for Video Generative Models''}
\titlerunning{Supplementary Material for Think-with-Video Reasoning}
\author{Meng Luo et al.}
\authorrunning{M. Luo et al.}
\institute{}
\maketitle
\else

\title{From Evaluation to Enhancement: Benchmarking and Improving Think-with-Video Reasoning for Video Generative Models}

\titlerunning{Think-with-Video Reasoning for Video Generative Models}

\author{Meng Luo\inst{1,\ast,\dagger} \and
Yicheng Liu\inst{2,\ast,\dagger} \and
Jiahao Wang\inst{3,\ddagger} \and
Yuanxing Zhang\inst{3} \and
Xin Tao\inst{3} \and
Pengfei Wan\inst{3} \and
Kun Gai\inst{3} \and
Hao Fei\inst{4,\ddagger}}

\authorrunning{M.~Luo et al.}

\institute{School of Computing, National University of Singapore, Singapore, Singapore\\
\and
Department of Computer Science and Technology,
Nanjing University, Nanjing, China
\and
Kling Team, Kuaishou Technology, Beijing, China
\and
Department of Computer Science, University of Oxford, Oxford, United Kingdom}

\maketitle
\begingroup
\renewcommand{\thefootnote}{\ifcase\value{footnote}\or\ensuremath{\ast}\or\ensuremath{\dagger}\or\ensuremath{\ddagger}\fi}
\footnotetext[1]{Equal contribution.}
\footnotetext[2]{Work done during an internship at Kling Team, Kuaishou Technology.}
\footnotetext[3]{Corresponding authors.}
\endgroup

\begin{abstract}
Video generation has advanced to produce visually compelling and temporally coherent results. Yet, whether these models can genuinely think with video---executing symbolic rules, respecting physical laws, and pursuing intentional goals---remains an open question. Existing benchmarks only partially address this, often conflating visual quality with cognitive correctness. We introduce \textbf{\texttt{VWG-Bench}} (\textbf{V}ideo \textbf{W}orld \textbf{G}eneralist Benchmark), a comprehensive benchmark spanning 9 reasoning dimensions and 38 fine-grained tasks. To enable precise diagnosis, we design a three-level VLM-as-Judge protocol that independently assesses video-level fluency, task-level rule adherence, and sample-level goal realization. Evaluations of leading models reveal a striking gap: while models achieve strong rendering scores, they consistently fail on logic-heavy and rule-constrained tasks. To address this, we propose \textbf{\texttt{Vid-PRE}} (\textbf{Vid}eo \textbf{P}rompt \textbf{R}easoner and \textbf{E}nhancer), a model-agnostic prompt rewriter that offloads the cognitive burden of reasoning to a dedicated VLM. Trained via reinforcement learning with purely text-based rewards, Vid-PRE produces concise, constraint-aware prompts without the instability of video-level reward signals. Experiments show that Vid-PRE yields substantial reasoning improvements across multiple generators without architectural modifications. Together, \texttt{VWG-Bench} and \texttt{Vid-PRE} offer a rigorous diagnostic lens and a scalable path toward true think-with-video capabilities. All data and code are publicly available at \url{https://huggingface.co/datasets/KlingTeam/VWG-Bench}.
\keywords{Video Generation \and Evaluation Benchmark \and Prompt Enhancement
\and Reinforcement Learning}
\end{abstract}

\input{sections/1_introduction}

\input{sections/2_benchmark}

\input{sections/3_vid_pre}

\input{sections/4_exp}

\input{sections/5_related_work}

\input{sections/6_conclusion_future_work}

% ---- Bibliography ----
\bibliographystyle{splncs04}
\bibliography{main}

\fi
\ifdefined\SUPPLEMENTONLY

% ---- Appendix ----
\appendix

\input{sections/supp_additional_evaluation_details}

\section{Task Descriptions}

\noindent\textbf{\color{blue}Task 1: Arithmetic}

\noindent\textbf{Definition:} Given a set of arithmetic expressions, the model is required to compute their results and write them in the corresponding positions.

\begin{figure}[htbp]
  \centering
  \includegraphics[width=0.50\linewidth]{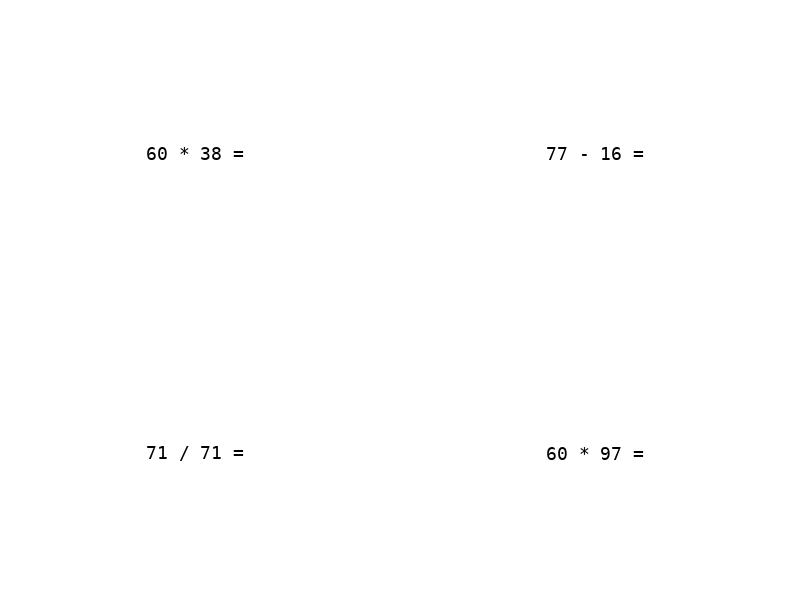}
  \caption{First-frame input used for I2V generation. \textbf{\textit{Prompt}}: ``Watch the equations appearing in this video. What is the value of $x$ in the last equation?''}
\end{figure}

\noindent\textbf{\color{blue}Task 2: Bubble Sorting}

\noindent\textbf{Definition:} Given a set of bubbles labeled with numbers, the model is required to make them rise one by one in ascending or descending order according to the numbers on the bubbles.

\begin{figure}[htbp]
  \centering
  \includegraphics[width=0.50\linewidth]{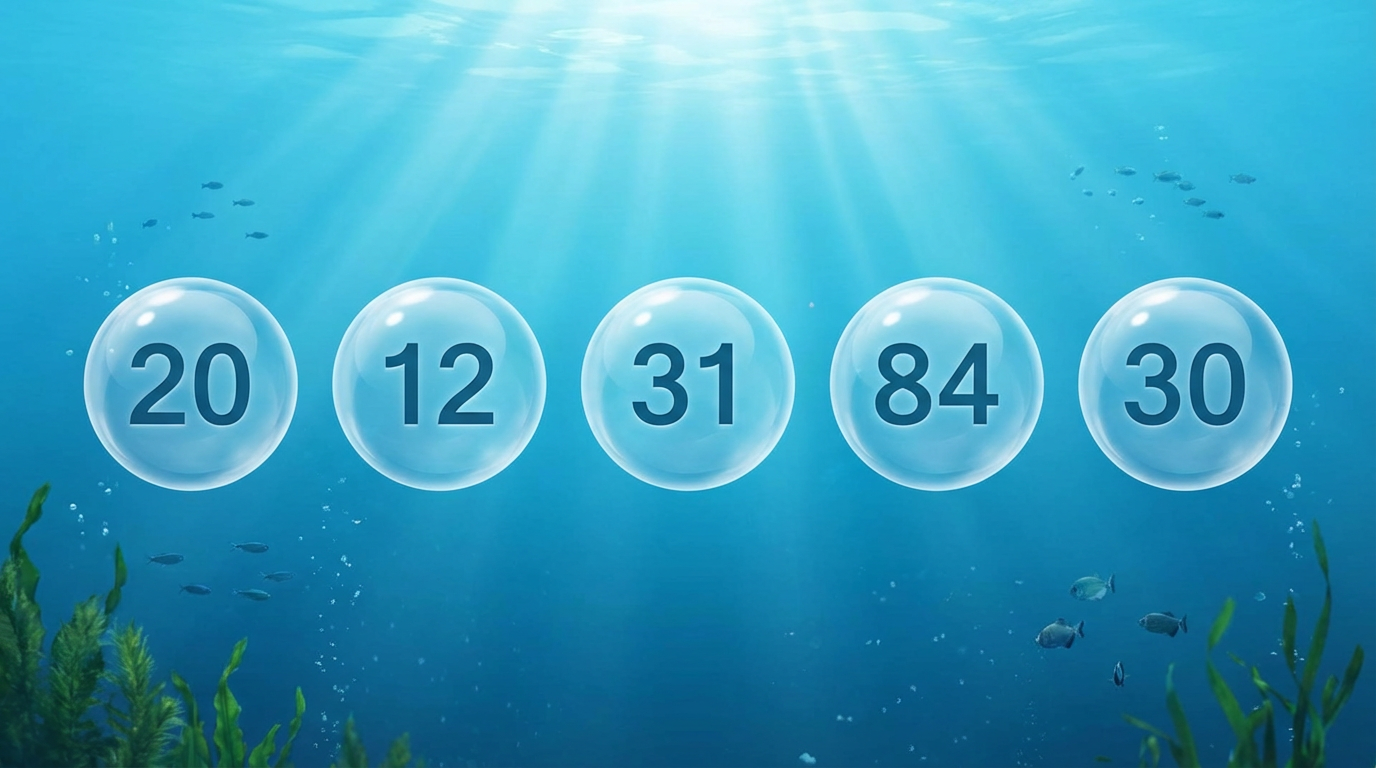}
  \caption{First-frame input used for I2V generation. \textbf{\textit{Prompt}}: ``Make the bubbles float out of the frame in ascending order according to their corresponding numbers.''}
\end{figure}

\clearpage
\noindent\textbf{\color{blue}Task 3: Coding}

\noindent\textbf{Definition:} Given a coding problem and an input, the model is required to compute the output and write it in the corresponding position.

\begin{figure}[htbp]
  \centering
  \includegraphics[width=0.50\linewidth]{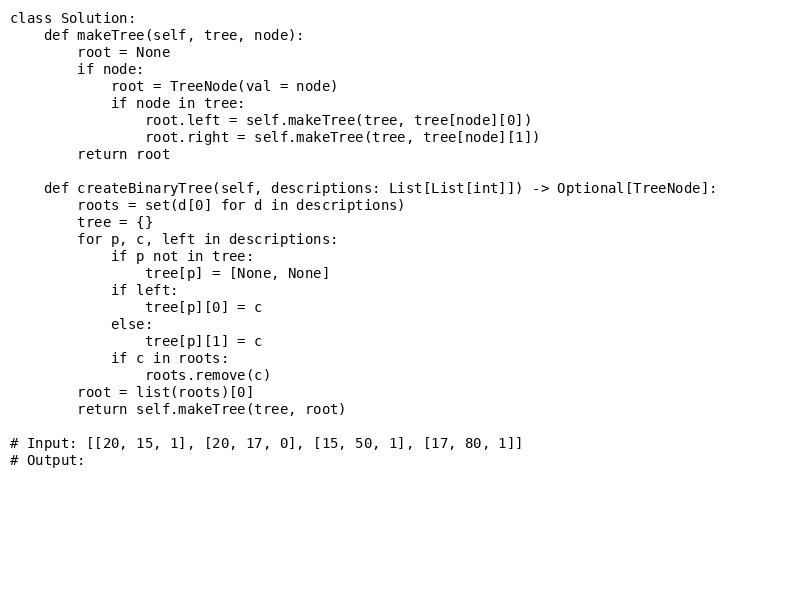}
  \caption{First-frame input used for I2V generation. \textbf{\textit{Prompt}}: ``Write the correct answer after \"Output\".''}
\end{figure}

\noindent\textbf{\color{blue}Task 4: Sudoku}

\noindent\textbf{Definition:} Given an incomplete Sudoku puzzle, the model is required to fill in the puzzle correctly.

\begin{figure}[htbp]
  \centering
  \includegraphics[width=0.50\linewidth]{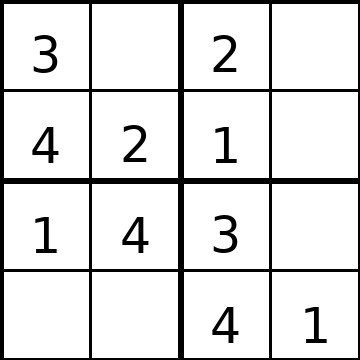}
  \caption{First-frame input used for I2V generation. \textbf{\textit{Prompt}}: ``Fill in this Sudoku puzzle.''}
\end{figure}

\clearpage
\noindent\textbf{\color{blue}Task 5: Tic-Tac-Toe}

\noindent\textbf{Definition:} Given a Tic-Tac-Toe board and several pieces, the model is required, as either ``X'' or ``O'', to make the next move that can either secure an immediate win or block the opponent's immediate win.

\begin{figure}[htbp]
  \centering
  \includegraphics[width=0.50\linewidth]{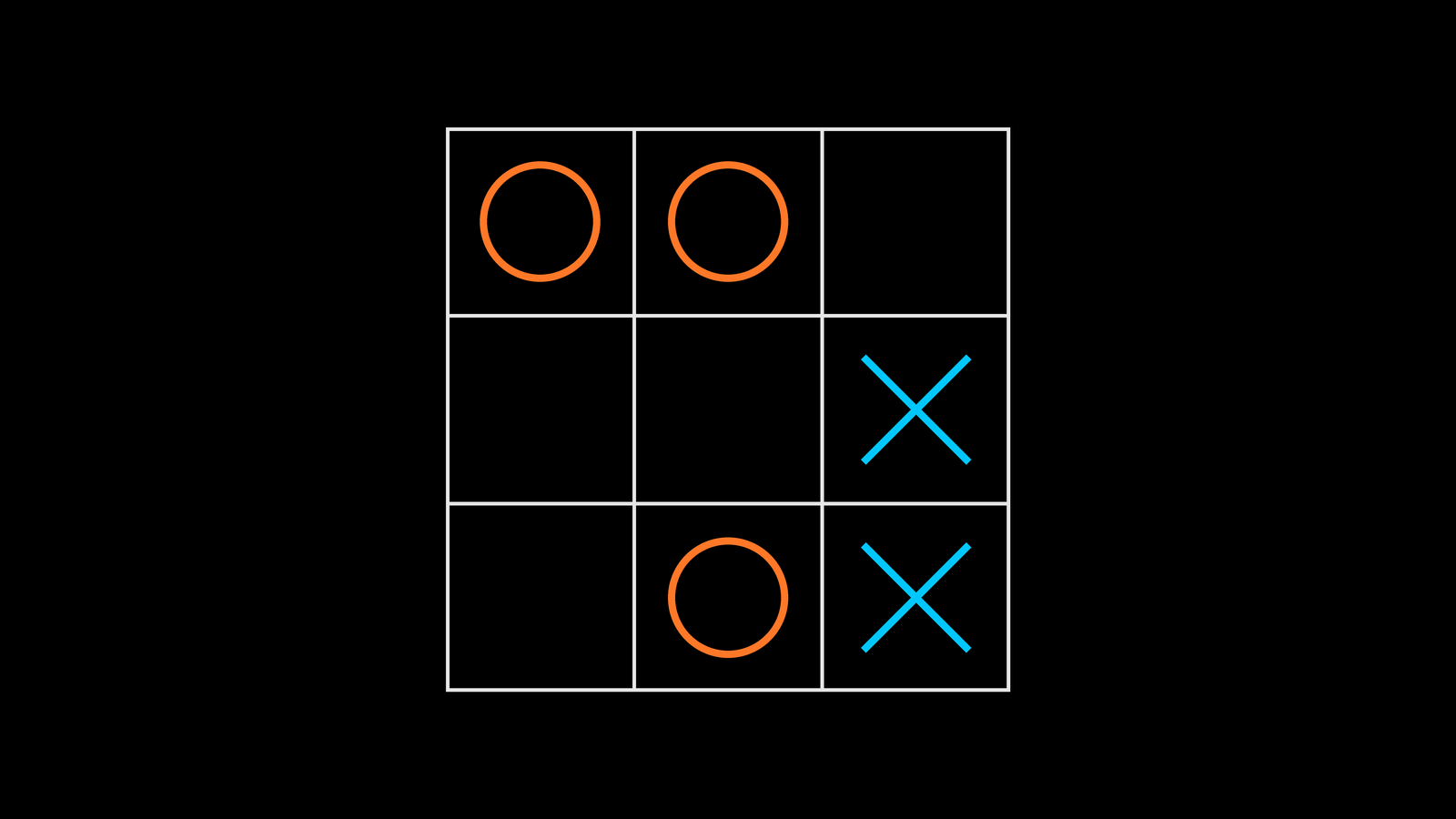}
  \caption{First-frame input used for I2V generation. \textbf{\textit{Prompt}}: ``Given the current state of the Tic-Tac-Toe board in the image, predict the next move that results in a win for the player X. Output the position of the winning move.''}
\end{figure}

\noindent\textbf{\color{blue}Task 6: Box Matching}

\noindent\textbf{Definition:} Given a set of objects and several boxes marked with certain attributes, the model is required to place each object into the corresponding box according to its attribute.

\begin{figure}[htbp]
  \centering
  \includegraphics[width=0.50\linewidth]{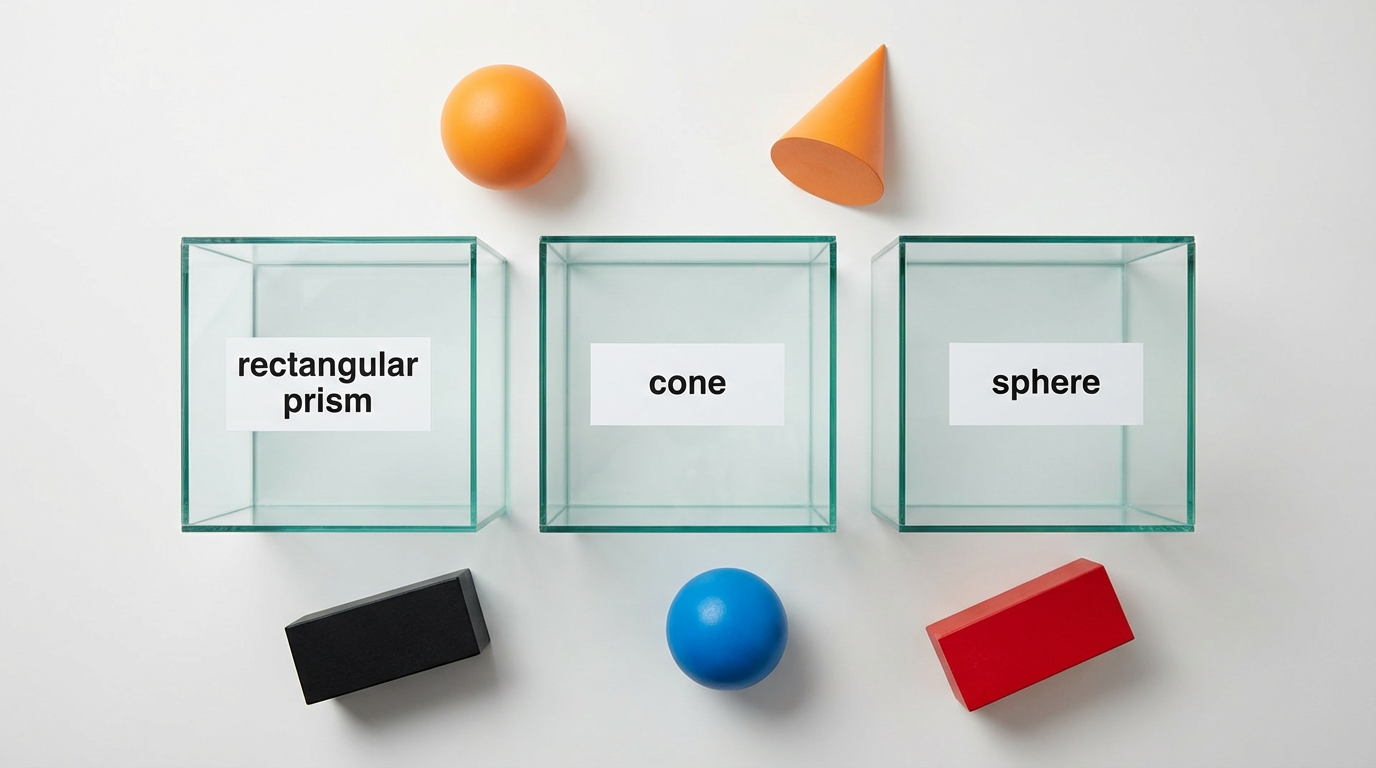}
  \caption{First-frame input used for I2V generation. \textbf{\textit{Prompt}}: ``Place the objects into the corresponding boxes according to the categories labeled on the boxes.''}
\end{figure}

\clearpage
\noindent\textbf{\color{blue}Task 7: Color Connecting}

\noindent\textbf{Definition:} Given several colored circles, the model is required to draw lines connecting circles of the same color.

\begin{figure}[htbp]
  \centering
  \includegraphics[width=0.50\linewidth]{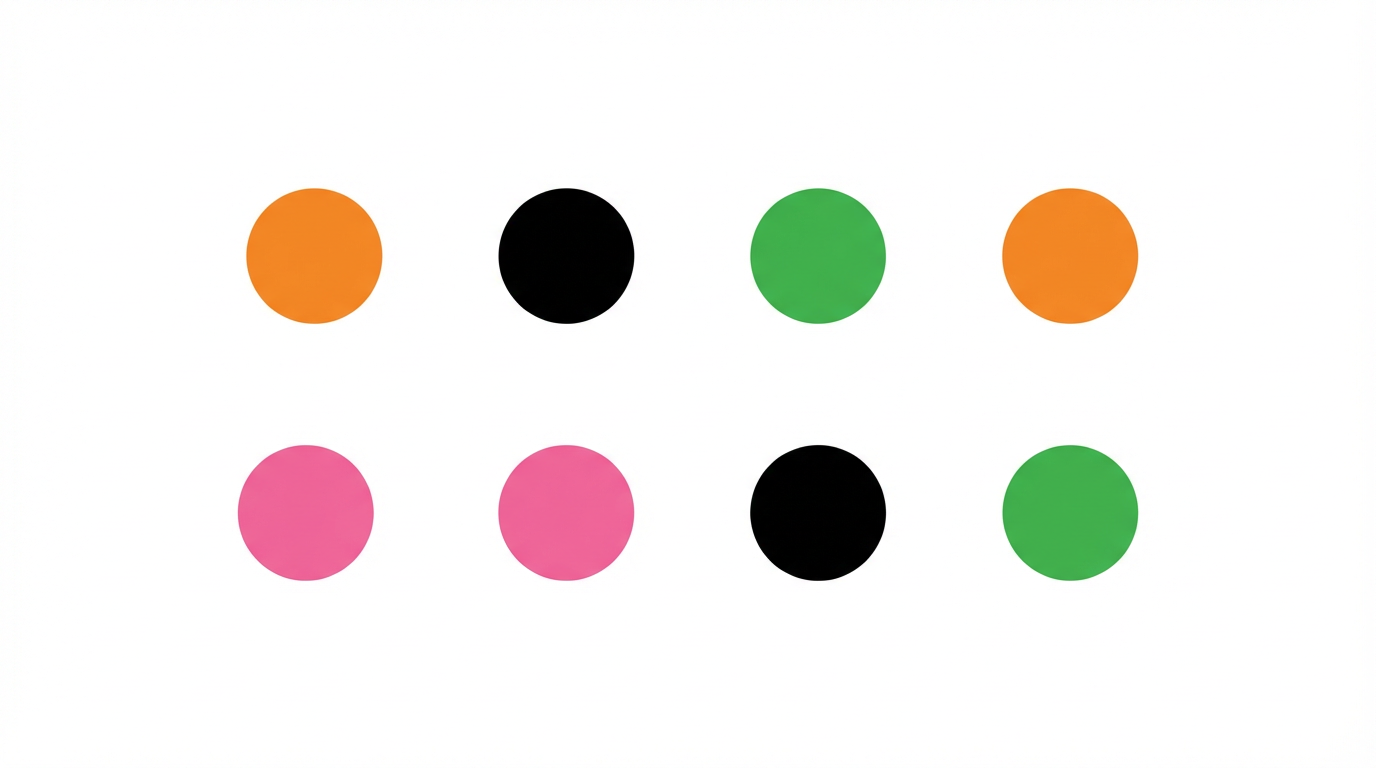}
  \caption{First-frame input used for I2V generation. \textbf{\textit{Prompt}}: ``Connect circles of the same color with lines, and ensure that the lines have no intersections or overlaps.''}
\end{figure}

\noindent\textbf{\color{blue}Task 8: Sequence Completion}

\noindent\textbf{Definition:} Given a sequence of continuous images and several blank slots, the model is required to infer the underlying pattern from the image sequence and fill the blank slots with the correct images.

\begin{figure}[htbp]
  \centering
  \includegraphics[width=0.50\linewidth]{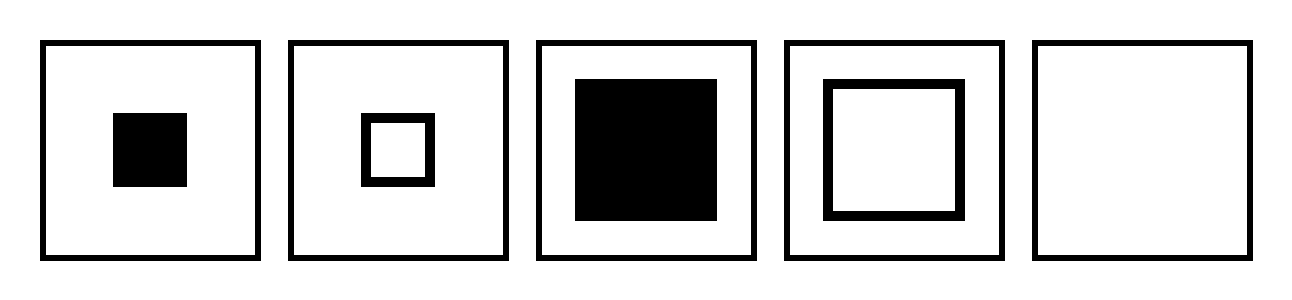}
  \caption{First-frame input used for I2V generation. \textbf{\textit{Prompt}}: ``Finish the incomplete frames of this shape-sequence puzzle. Do not change given patterns. Each row contains equal-sized square cells outlined in black on white. Shapes are solid black arrows, dots, or stripes that obey a clear mathematical rule (rotation, translation, scaling, or toggling). Produce the missing panels so the sequence remains consistent and has a single logical solution.''}
\end{figure}

\clearpage
\noindent\textbf{\color{blue}Task 9: Shape Fitting}

\noindent\textbf{Definition:} Given several 3D shapes and a board with multiple shape-shaped slots, the model is required to place each 3D shape into the corresponding slot based on its shape.

\begin{figure}[htbp]
  \centering
  \includegraphics[width=0.50\linewidth]{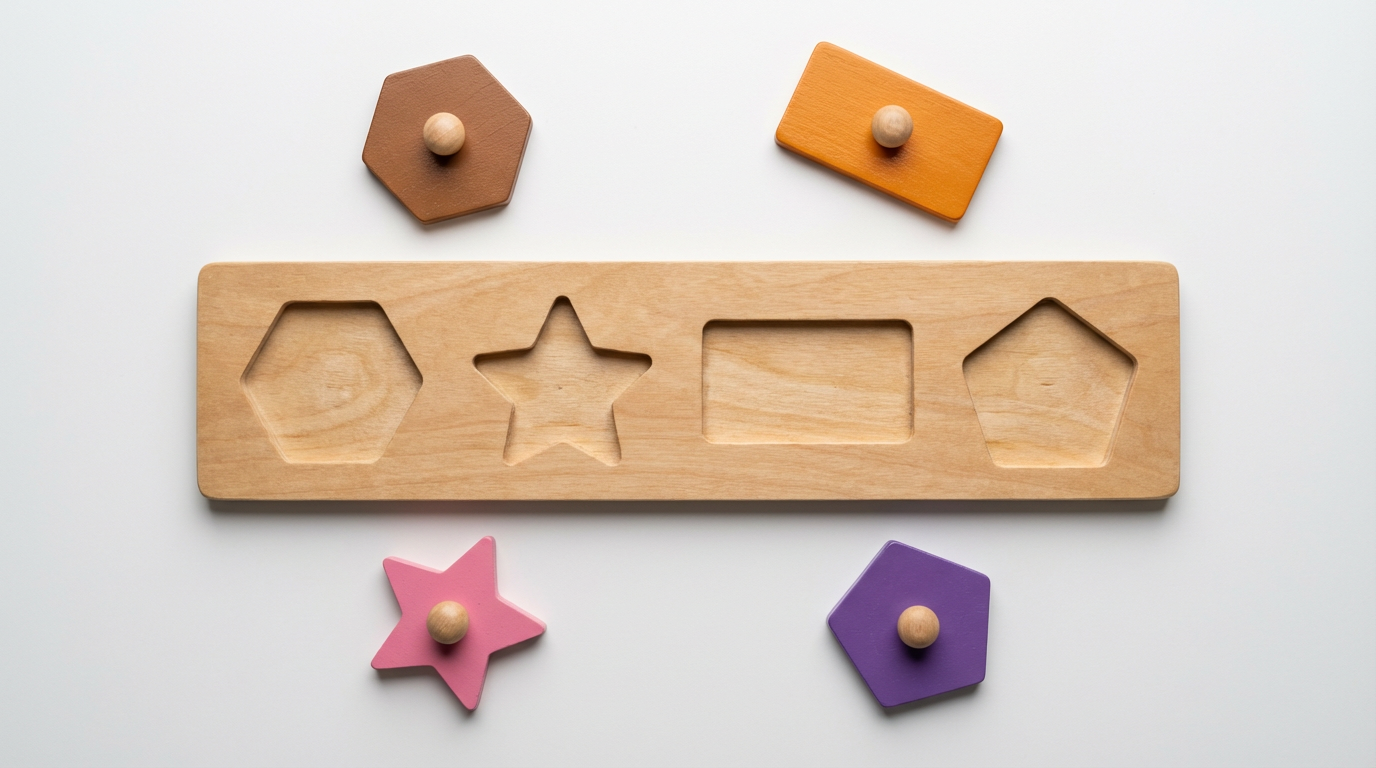}
  \caption{First-frame input used for I2V generation. \textbf{\textit{Prompt}}: ``The scene shows some colored pieces, and a panel with some holes. Each colored piece fits into one and only one hole. A hand grabs each colored piece and puts it into an empty hole that has the exact same shape.''}
\end{figure}

\noindent\textbf{\color{blue}Task 10: Visual Analogy}

\noindent\textbf{Definition:} Given a row of patterns and the patterns after a rule-based transformation, the model is required to apply the same transformation rule to a new row of patterns and produce the result.

\begin{figure}[htbp]
  \centering
  \includegraphics[width=0.50\linewidth]{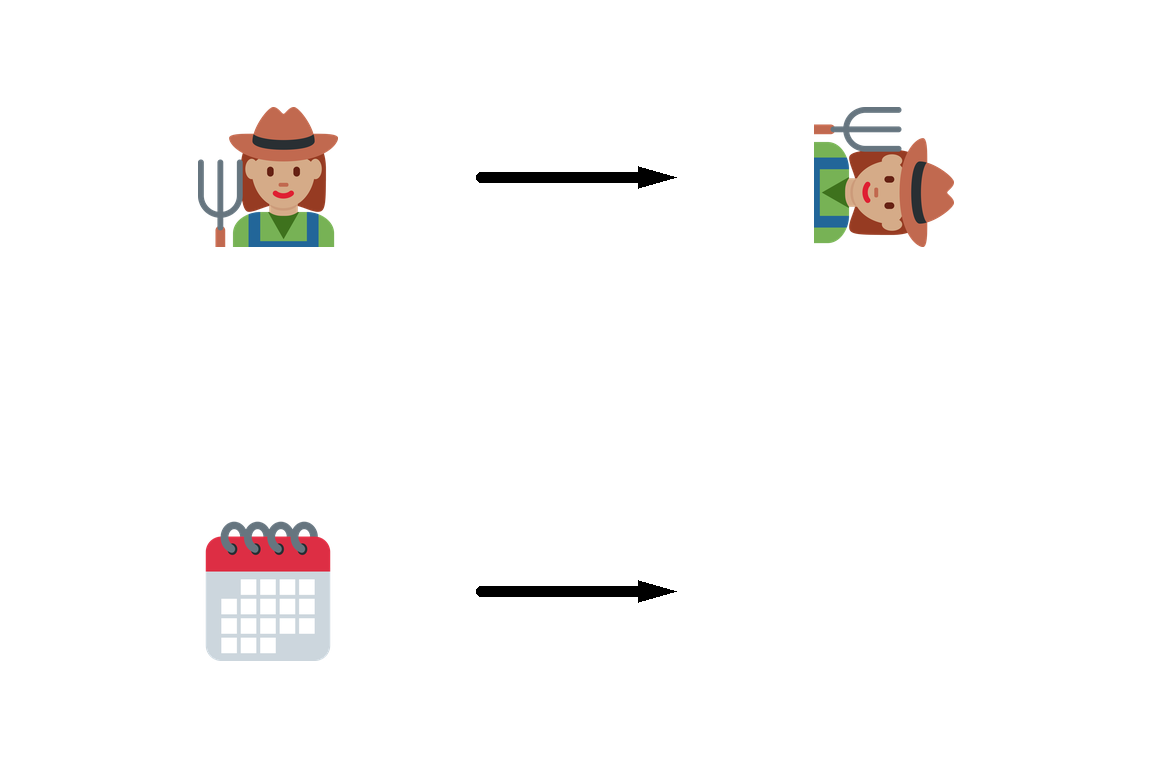}
  \caption{First-frame input used for I2V generation. \textbf{\textit{Prompt}}: ``Create a smooth animation to generate the missing object in the lower right region and solve the visual analogy. The original three objects must remain still. Static shot, no zoom no pan no dolly.''}
\end{figure}

\clearpage
\noindent\textbf{\color{blue}Task 11: Embodied Action Execution}

\noindent\textbf{Definition:} Given an embodied scene and an action instruction, the model is required to control a robotic arm or a human hand to execute the action.

\begin{figure}[htbp]
  \centering
  \includegraphics[width=0.50\linewidth]{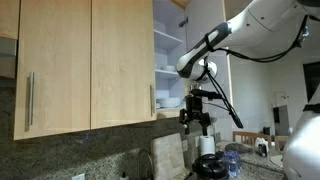}
  \caption{First-frame input used for I2V generation. \textbf{\textit{Prompt}}: ``From a first-person perspective, execute this instruction with human/robotic arm(s): closing the rightest cabinet door.''}
\end{figure}

\noindent\textbf{\color{blue}Task 12: Embodied Trajectory Reasoning}

\noindent\textbf{Definition:} Given an embodied scene, an action instruction, and several candidate trajectories, the model is required to control a robotic arm or a human hand to execute the action along one of the candidate trajectories.

\begin{figure}[htbp]
  \centering
  \includegraphics[width=0.50\linewidth]{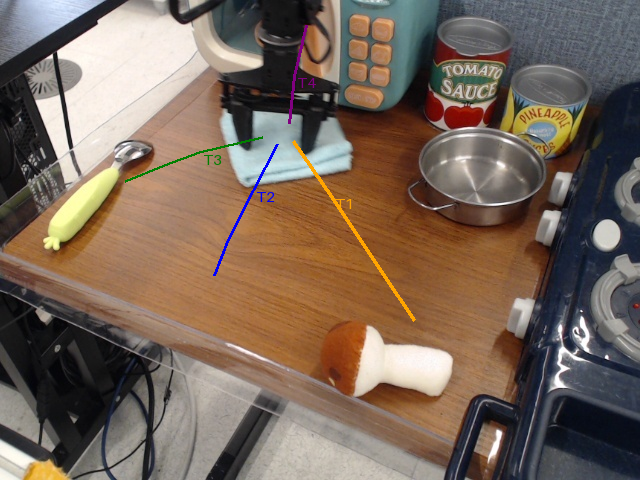}
  \caption{First-frame input used for I2V generation. \textbf{\textit{Prompt}}: ``From a first-person perspective, execute this instruction with human/robotic arm(s): drag the cloth to the middle of a spoon and a mushroom. There are four possible trajectories (T1 ~T4) shown in the image; please operate by following the trajectory you consider to be the most correct.''}
\end{figure}

\clearpage
\noindent\textbf{\color{blue}Task 13: Implicit Goal Reasoning}

\noindent\textbf{Definition:} Given an embodied scene and an implicit action goal, the model is required to infer and execute the true action instruction.

\begin{figure}[htbp]
  \centering
  \includegraphics[width=0.50\linewidth]{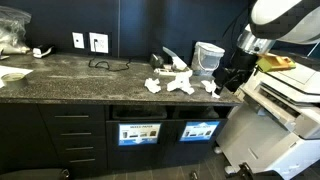}
  \caption{First-frame input used for I2V generation. \textbf{\textit{Prompt}}: ``From a first-person perspective, how do I operate the robotic arm to complete the instruction: \"help me clean up the tissues on the table\"?''}
\end{figure}

\noindent\textbf{\color{blue}Task 14: Intention Causality}

\noindent\textbf{Definition:} Given a scene and several visual cues, the model is required to infer the behavioral goal from the visual cues and execute it.

\begin{figure}[htbp]
  \centering
  \includegraphics[width=0.50\linewidth]{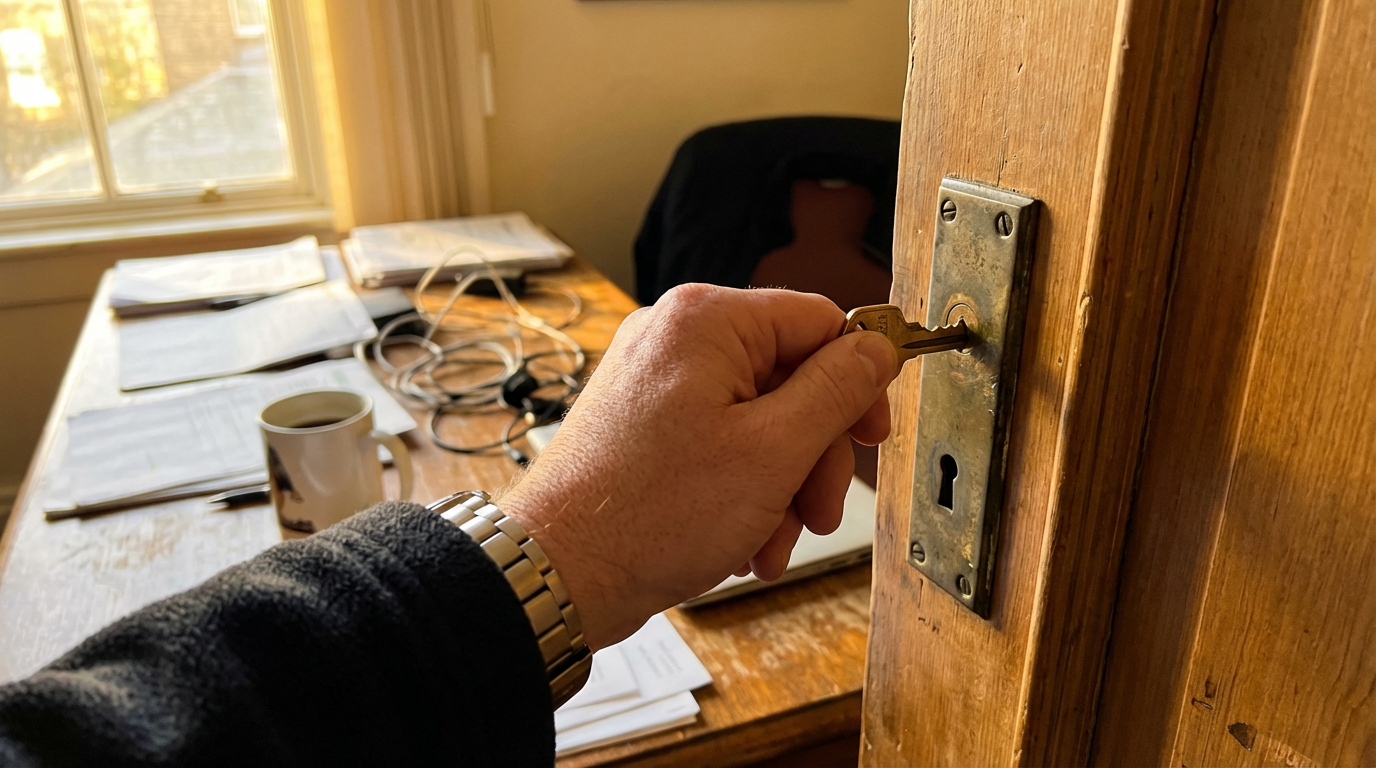}
  \caption{First-frame input used for I2V generation. \textbf{\textit{Prompt}}: ``Inserting the key into the keyhole.''}
\end{figure}

\clearpage
\noindent\textbf{\color{blue}Task 15: Procedural Knowledge}

\noindent\textbf{Definition:} Given a scene and a complex action instruction, the model is required to understand, decompose, and correctly execute the instruction.

\begin{figure}[htbp]
  \centering
  \includegraphics[width=0.50\linewidth]{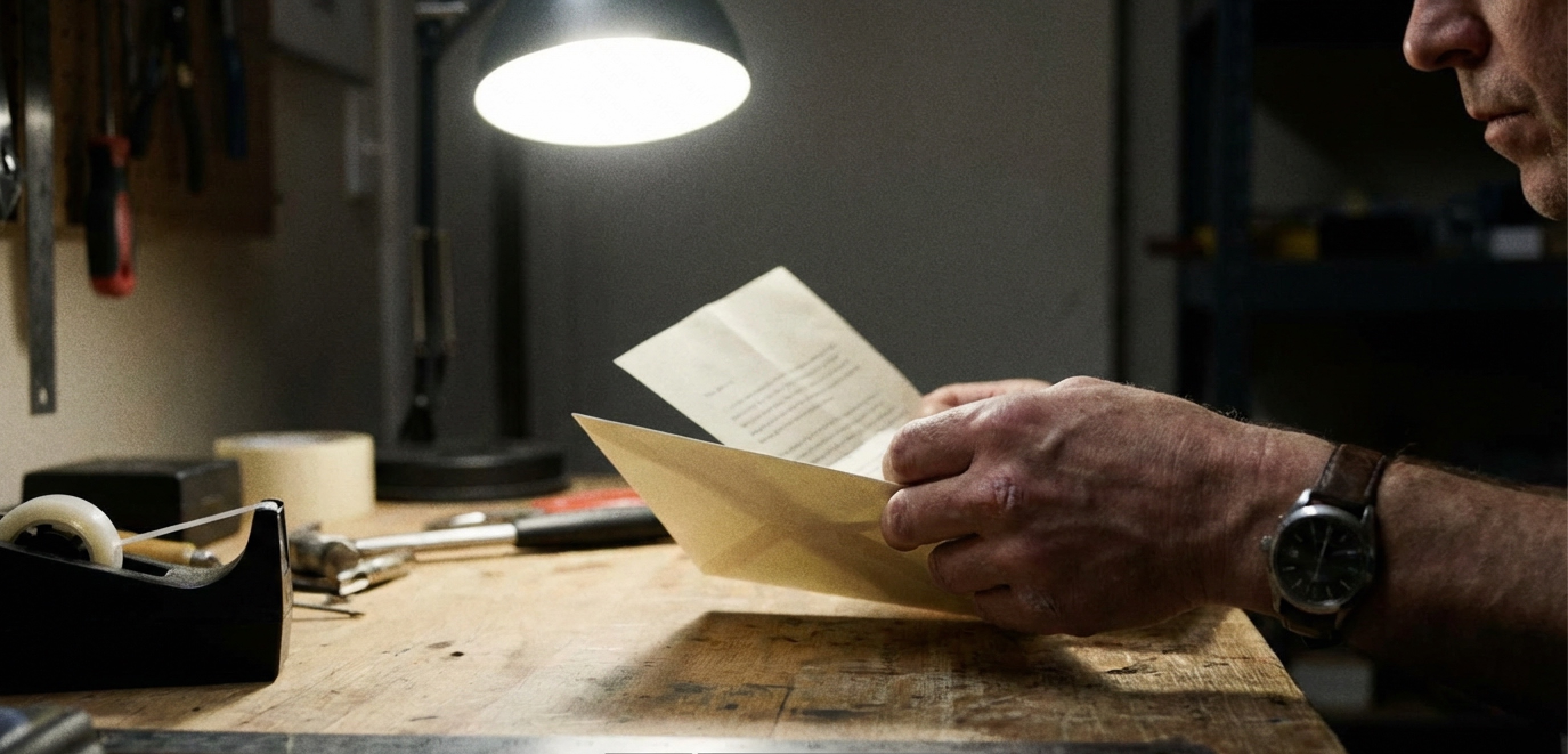}
  \caption{First-frame input used for I2V generation. \textbf{\textit{Prompt}}: ``Folding the letter, inserting it into the envelope, and sealing it.''}
\end{figure}

\noindent\textbf{\color{blue}Task 16: GUI Instruction Execution}

\noindent\textbf{Definition:} Given a GUI interface and an instruction, the model is required to perform the operation and correctly present the subsequent changes in the interface.

\begin{figure}[htbp]
  \centering
  \includegraphics[width=0.50\linewidth]{figs/GUI_Instruction_Execution.png}
  \caption{First-frame input used for I2V generation. \textbf{\textit{Prompt}}: ``Execute this instruction: Open the ParEval Learderboard.''}
\end{figure}

\clearpage
\noindent\textbf{\color{blue}Task 17: GUI Click Reasoning}

\noindent\textbf{Definition:} Given a GUI interface and a specific location, the model is required to execute a click at that location and plausibly present the subsequent changes in the interface.

\begin{figure}[htbp]
  \centering
  \includegraphics[width=0.25\linewidth]{figs/GUI_Click_Reasoning.png}
  \caption{First-frame input used for I2V generation. \textbf{\textit{Prompt}}: ``Click \"A prominent play button labeled 'Latest Episode' displayed in the middle section of the podcast page\".''}
\end{figure}

\noindent\textbf{\color{blue}Task 18: Physics Inversion}

\noindent\textbf{Definition:} Given a standard physical scene, the model is required to simulate a counterintuitive physical evolution process in which fundamental physical laws are reversed.

\begin{figure}[htbp]
  \centering
  \includegraphics[width=0.50\linewidth]{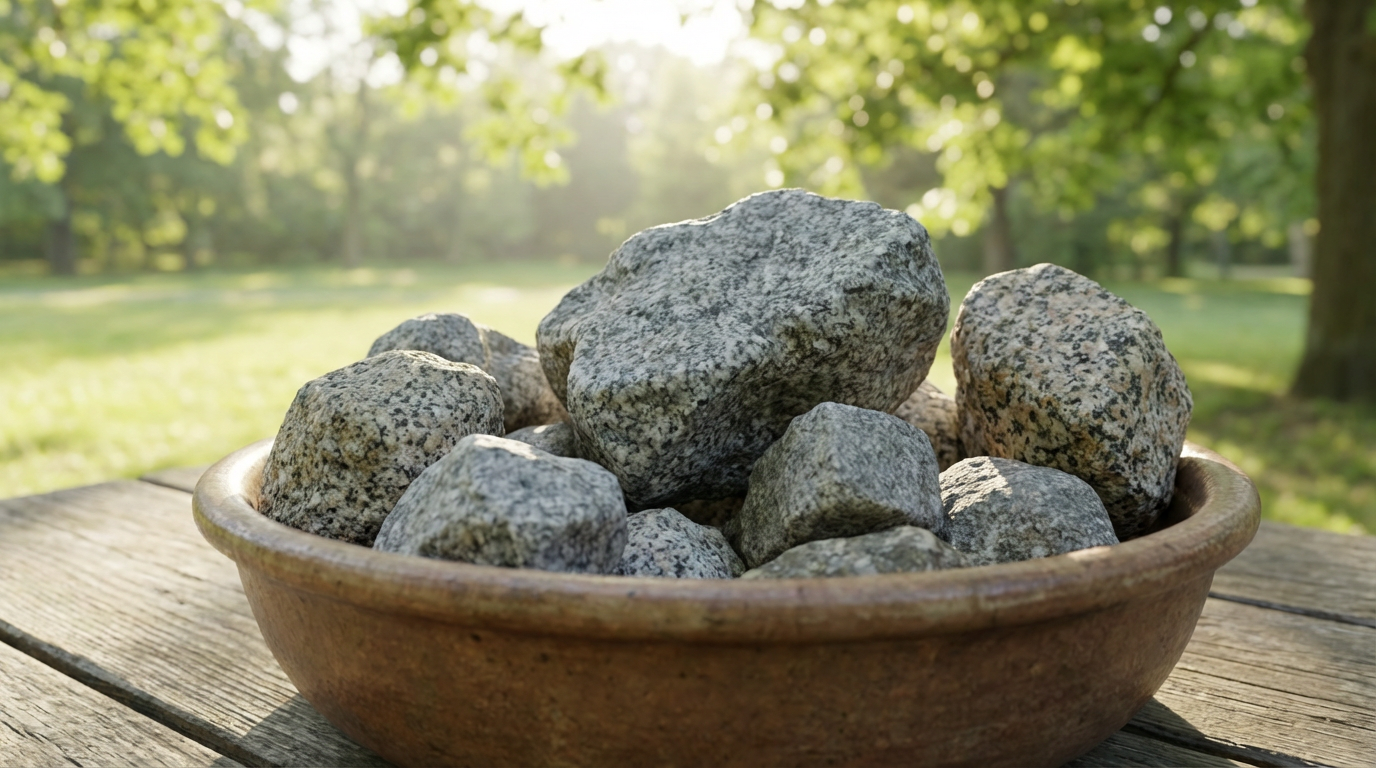}
  \caption{First-frame input used for I2V generation. \textbf{\textit{Prompt}}: ``The solid granite rocks gently lift off the table and float upwards into the sky.''}
\end{figure}

\clearpage
\noindent\textbf{\color{blue}Task 19: Causal Intervention}

\noindent\textbf{Definition:} Given a causal sequence, the model is required to alter one key intervention variable, infer the resulting downstream chain reactions, and generate the subsequent outcomes.

\begin{figure}[htbp]
  \centering
  \includegraphics[width=0.50\linewidth]{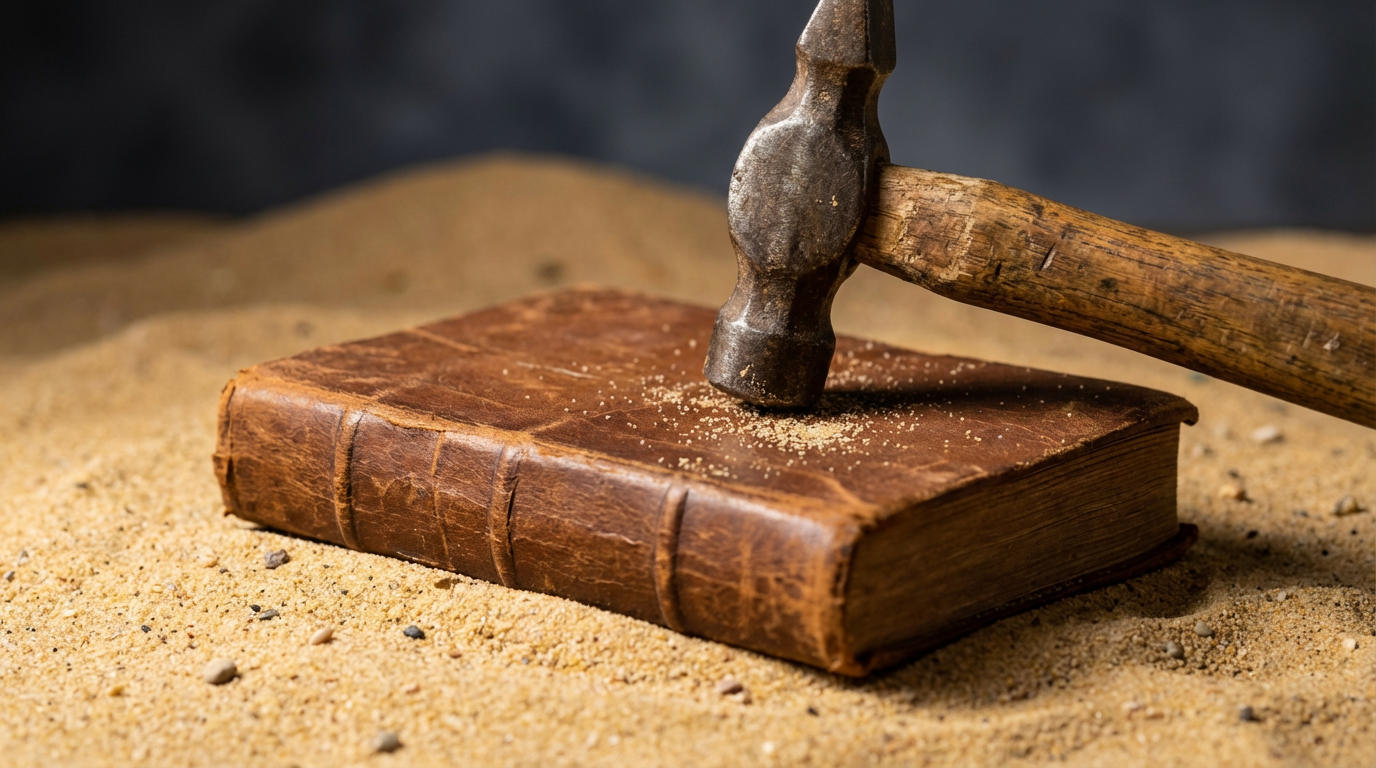}
  \caption{First-frame input used for I2V generation. \textbf{\textit{Prompt}}: ``The hammer hits the book hard. The book must remain completely still for several seconds after being hit. After the delay, the book suddenly flies away violently.''}
\end{figure}

\noindent\textbf{\color{blue}Task 20: Fictional Rule Execution}

\noindent\textbf{Definition:} Given a set of fictional operational rules, the model is required to drive objects through state transitions strictly according to those rules.

\begin{figure}[htbp]
  \centering
  \includegraphics[width=0.50\linewidth]{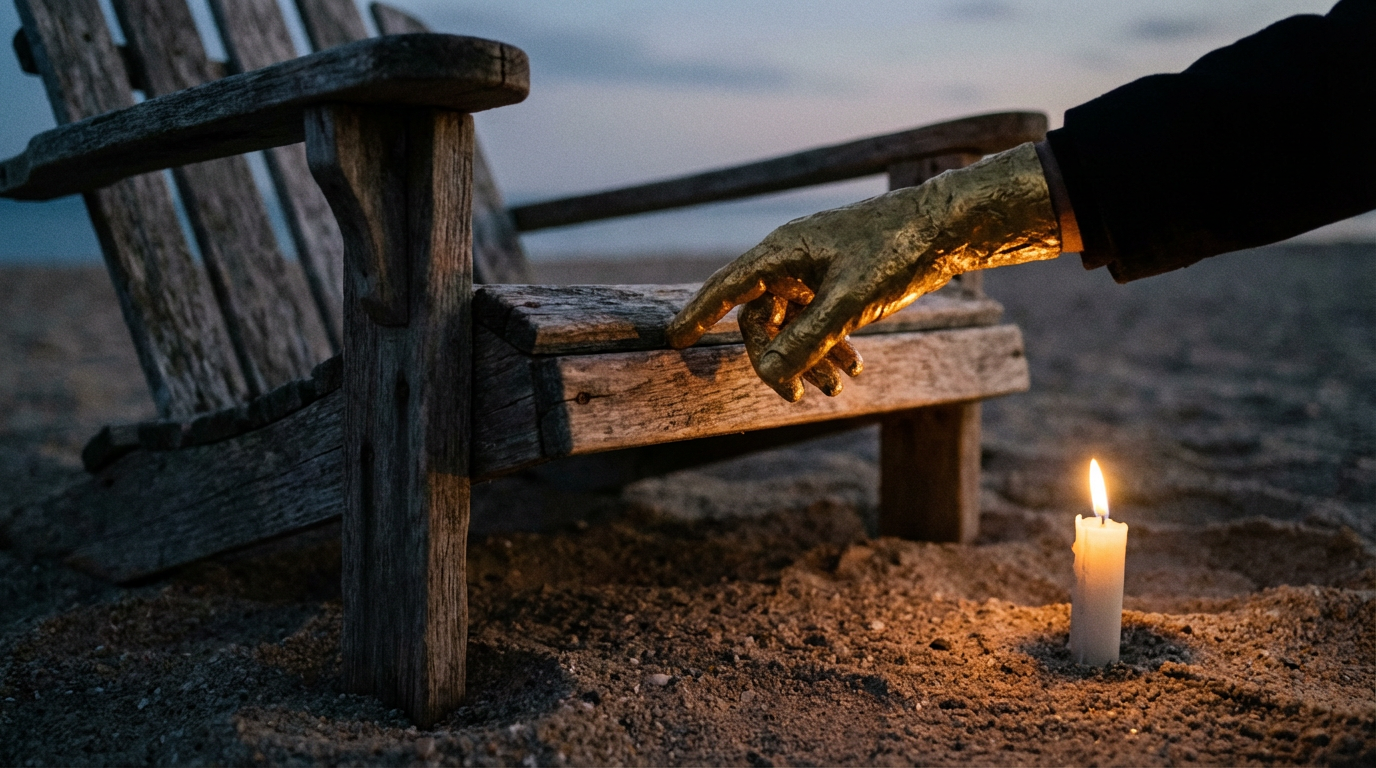}
  \caption{First-frame input used for I2V generation. \textbf{\textit{Prompt}}: ``The solid gold finger touches the wooden chair, transforming the entire chair into solid gold.''}
\end{figure}

\clearpage
\noindent\textbf{\color{blue}Task 21: Surreal Attribute Mapping}

\noindent\textbf{Definition:} Given objects with surreal properties, the model is required to simulate their anomalous physical behavior under specific forces.

\begin{figure}[htbp]
  \centering
  \includegraphics[width=0.50\linewidth]{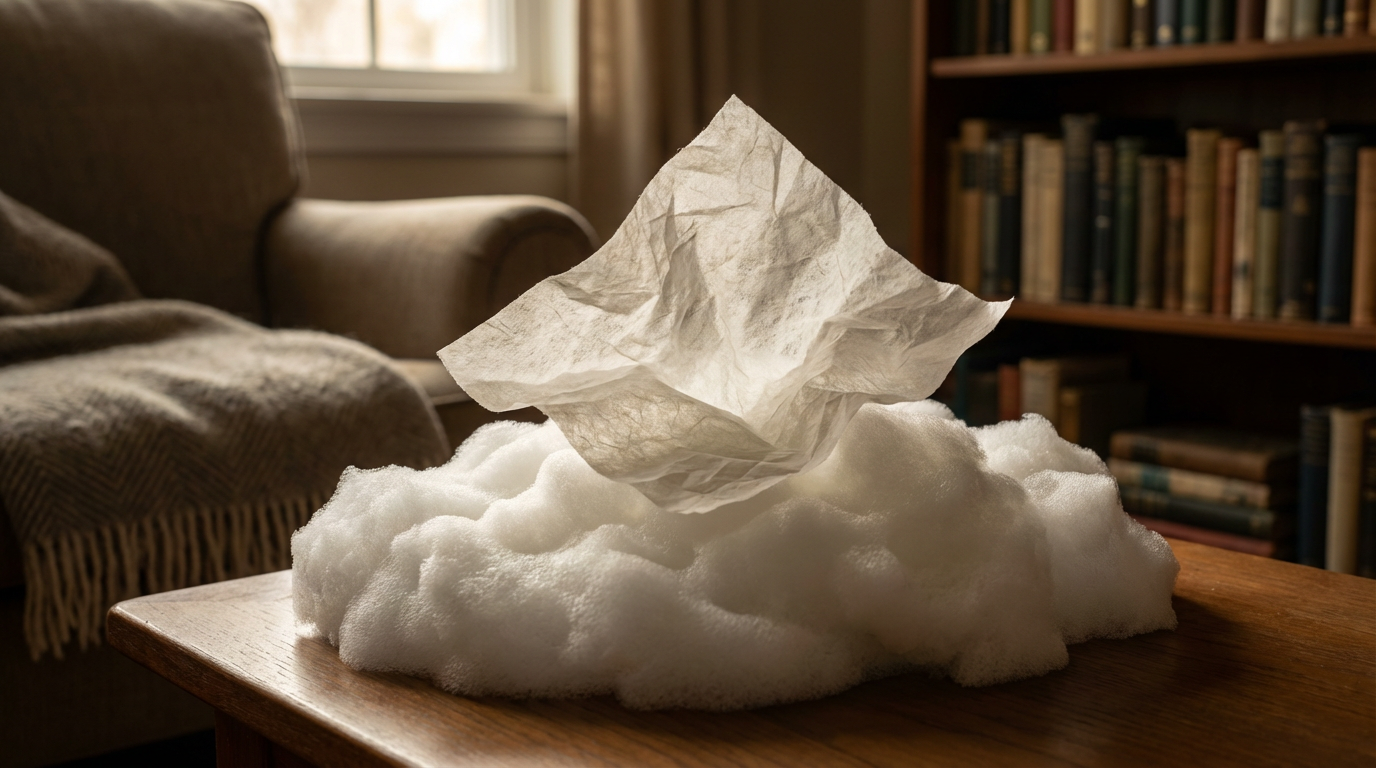}
  \caption{First-frame input used for I2V generation. \textbf{\textit{Prompt}}: ``The tissue paper is resting on the foam. Despite its light appearance, the leaf behaves as if it weighs 1000kg.''}
\end{figure}

\noindent\textbf{\color{blue}Task 22: Reverse Temporal Logic}

\noindent\textbf{Definition:} Given an object, the model is required to generate its physical evolution under reversed time.

\begin{figure}[htbp]
  \centering
  \includegraphics[width=0.50\linewidth]{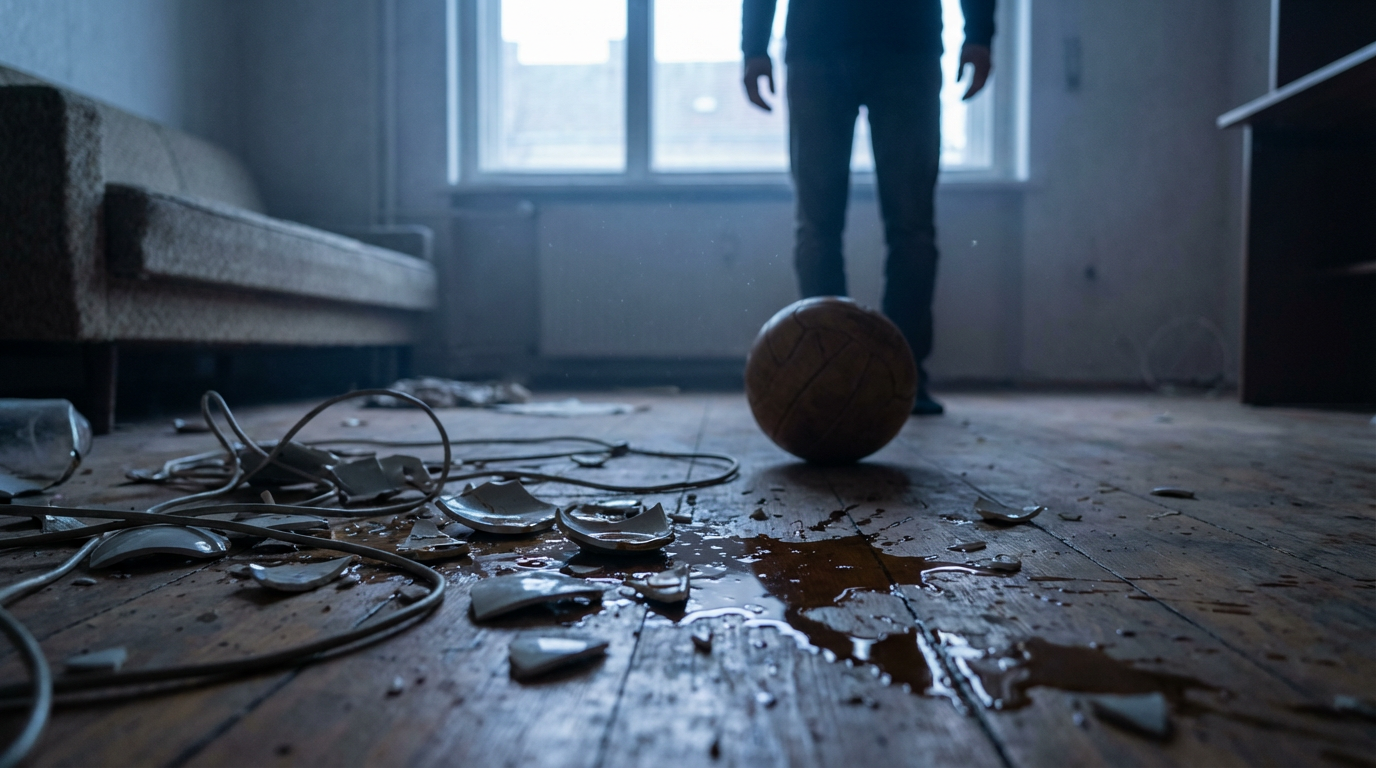}
  \caption{First-frame input used for I2V generation. \textbf{\textit{Prompt}}: ``The ball rolling backward, accelerating, and jumping precisely back into the person's hand.''}
\end{figure}

\clearpage
\noindent\textbf{\color{blue}Task 23: Maze}

\noindent\textbf{Definition:} Given a maze and a pair of start and end points, the model is required to move a block from the start point to the end point.

\begin{figure}[htbp]
  \centering
  \includegraphics[width=0.50\linewidth]{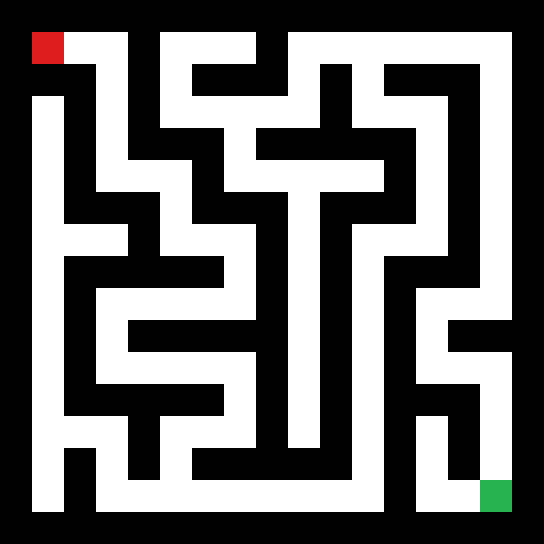}
  \caption{First-frame input used for I2V generation. \textbf{\textit{Prompt}}: ``Move the green cube along the maze to the red cube.''}
\end{figure}

\noindent\textbf{\color{blue}Task 24: Real World Spatial Reasoning}

\noindent\textbf{Definition:} Given a real-world scene, the model is required to complete instructions or reasoning involving real-world objects in the scene.

\begin{figure}[htbp]
  \centering
  \includegraphics[width=0.50\linewidth]{figs/Real_World_Spatial_Reasoning.png}
  \caption{First-frame input used for I2V generation. \textbf{\textit{Prompt}}: ``A blue arrow from the catcher in the red jersey toward the batter in the green jersey.''}
\end{figure}

\clearpage
\noindent\textbf{\color{blue}Task 25: Rotation Reasoning}

\noindent\textbf{Definition:} Given a rotated image, the model is required to rotate it back by a fixed angle and direction and use a bounding box to mark the corresponding object.

\begin{figure}[htbp]
  \centering
  \includegraphics[width=0.50\linewidth]{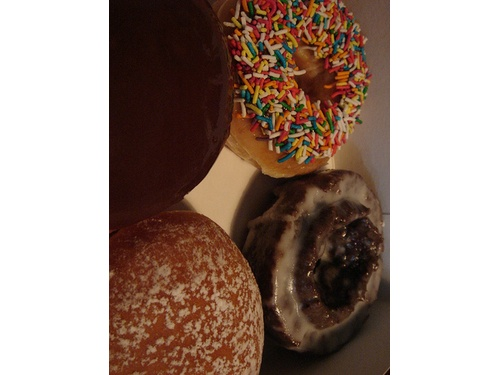}
  \caption{First-frame input used for I2V generation. \textbf{\textit{Prompt}}: ``Rotate the video frame 90 degrees clockwise in the 2D plane, then draw bounding boxes around donut.''}
\end{figure}

\noindent\textbf{\color{blue}Task 26: Spatial Geometry}

\noindent\textbf{Definition:} Given a 3D image, the model is required to apply the specified 3D transformation to the object in the image.

\begin{figure}[htbp]
  \centering
  \includegraphics[width=0.50\linewidth]{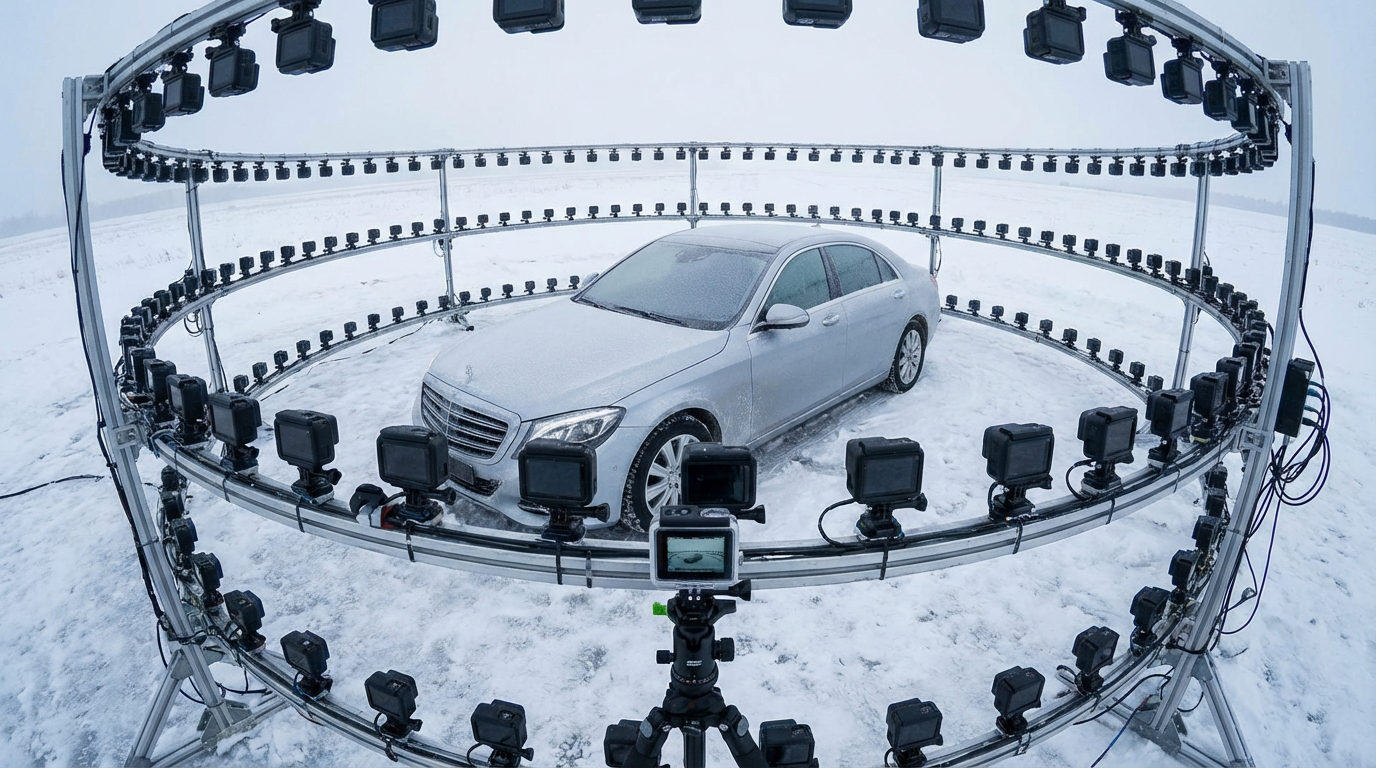}
  \caption{First-frame input used for I2V generation. \textbf{\textit{Prompt}}: ``The view switching smoothly around the still car.''}
\end{figure}

\clearpage
\noindent\textbf{\color{blue}Task 27: Short-term Temporal Dynamics}

\noindent\textbf{Definition:} Given a scene and several objects, the model is required to generate their state changes over the course of several seconds.

\begin{figure}[htbp]
  \centering
  \includegraphics[width=0.50\linewidth]{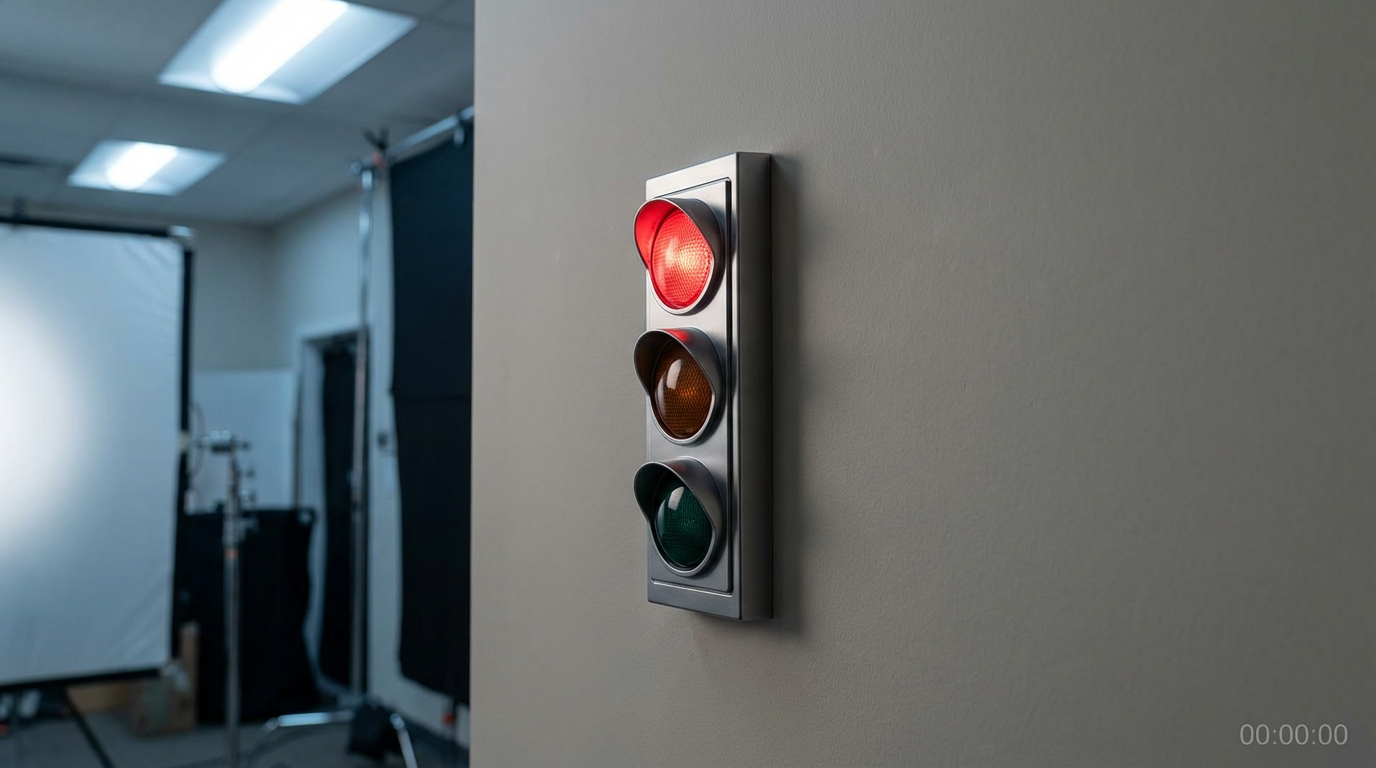}
  \caption{First-frame input used for I2V generation. \textbf{\textit{Prompt}}: ``The signal switching to green.''}
\end{figure}

\noindent\textbf{\color{blue}Task 28: Medium-term Evolutionary Processes}

\noindent\textbf{Definition:} Given a scene and several objects, the model is required to generate their state changes over timescales ranging from minutes to months.

\begin{figure}[htbp]
  \centering
  \includegraphics[width=0.50\linewidth]{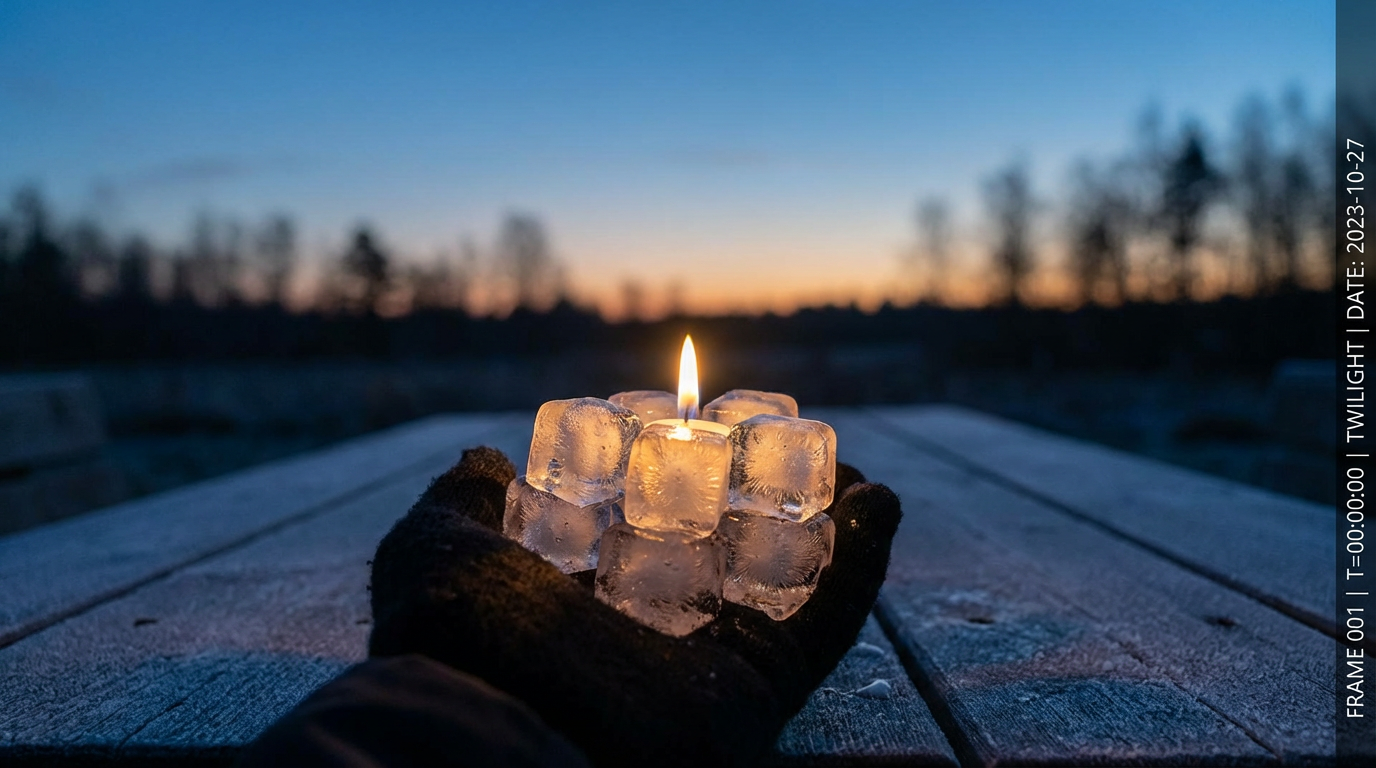}
  \caption{First-frame input used for I2V generation. \textbf{\textit{Prompt}}: ``A time-lapse of the candle burning over an hour.''}
\end{figure}

\clearpage
\noindent\textbf{\color{blue}Task 29: Long-term Temporal Progression}

\noindent\textbf{Definition:} Given a scene and several objects, the model is required to generate their long-term evolution on a yearly timescale.

\begin{figure}[htbp]
  \centering
  \includegraphics[width=0.50\linewidth]{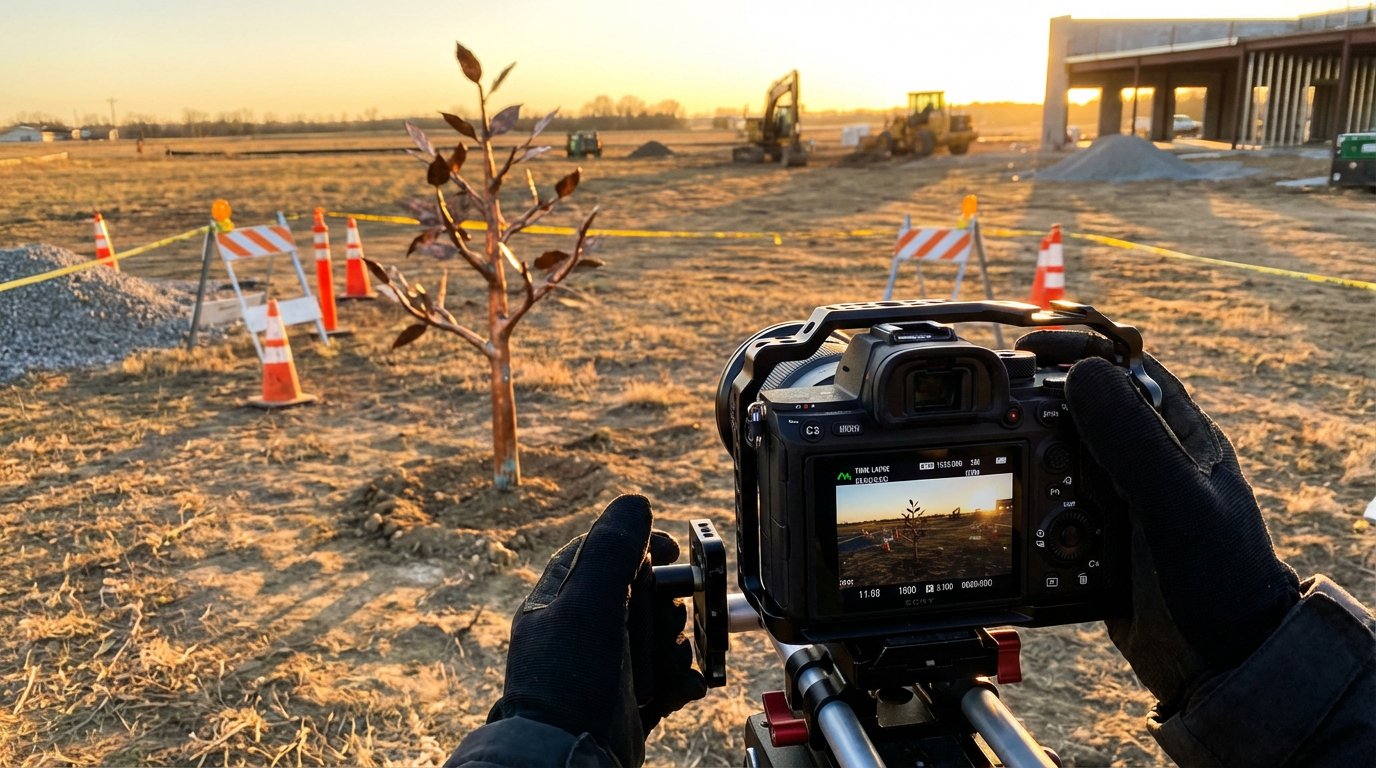}
  \caption{First-frame input used for I2V generation. \textbf{\textit{Prompt}}: ``A time-lapse over five years.''}
\end{figure}

\noindent\textbf{\color{blue}Task 30: Physics-based Reasoning}

\noindent\textbf{Definition:} Given a physical scene, the model is required to generate physics-related changes of objects in the scene, involving mechanics and dynamics, fluid dynamics, material deformation, optics, thermodynamics, and acoustics.

\begin{figure}[htbp]
  \centering
  \includegraphics[width=0.50\linewidth]{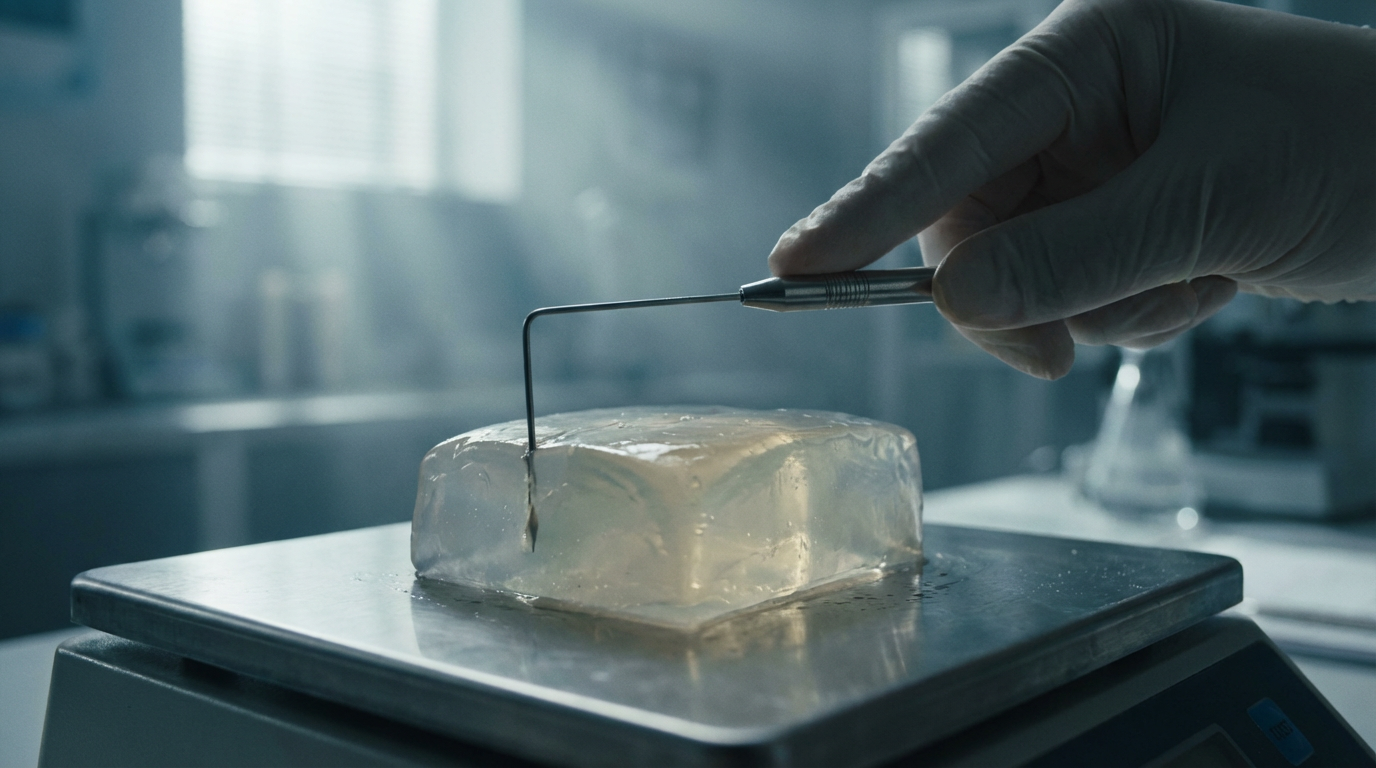}
  \caption{First-frame input used for I2V generation. \textbf{\textit{Prompt}}: ``The knife cutting through the block.''}
\end{figure}

\clearpage
\noindent\textbf{\color{blue}Task 31: Biological Kinematics}

\noindent\textbf{Definition:} Given a biological-entity scene, the model is required to generate its natural life motion while adhering to physiological structure and kinematic constraints.

\begin{figure}[htbp]
  \centering
  \includegraphics[width=0.50\linewidth]{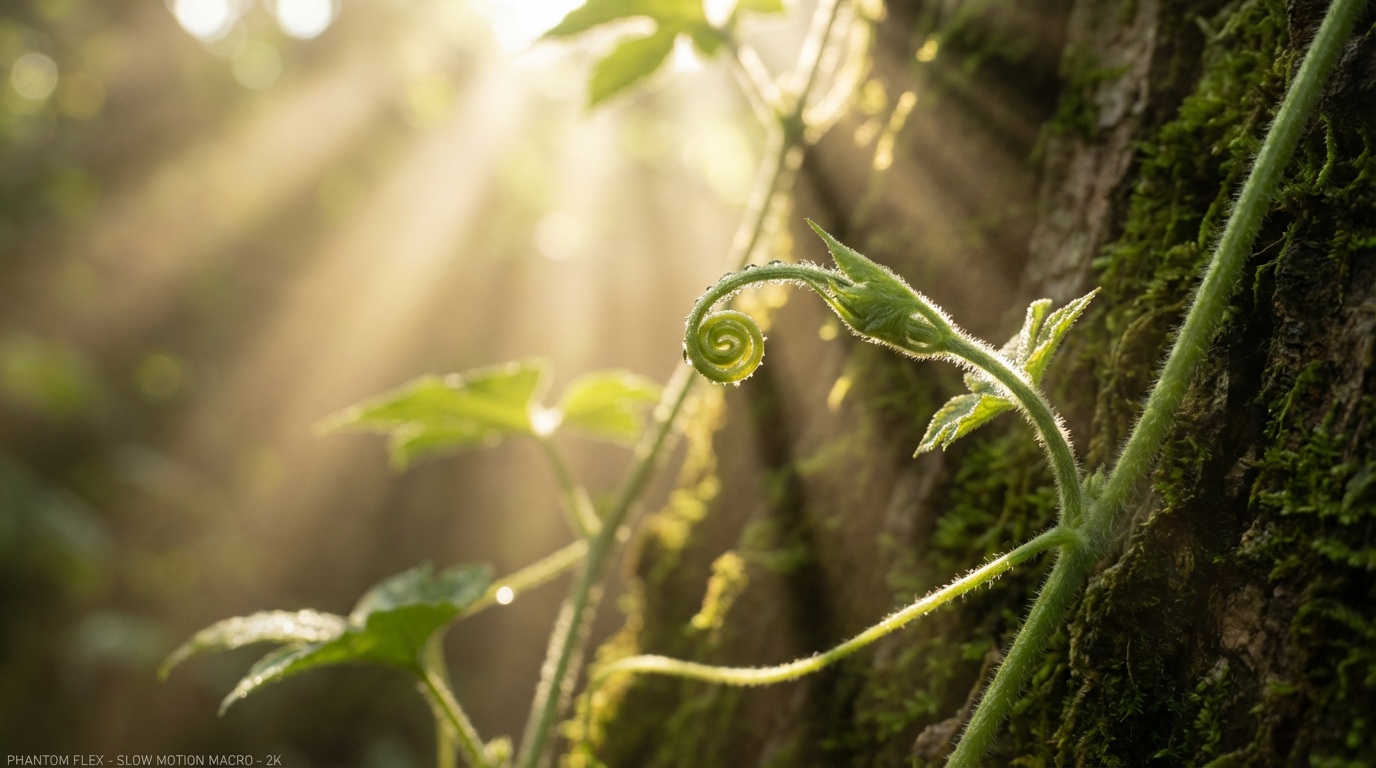}
  \caption{First-frame input used for I2V generation. \textbf{\textit{Prompt}}: ``The plant growing towards light.''}
\end{figure}

\noindent\textbf{\color{blue}Task 32: Physical Commonsense Reasoning}

\noindent\textbf{Definition:} Given a scene, the model is required to generate physical changes governed by everyday physical common sense.

\begin{figure}[htbp]
  \centering
  \includegraphics[width=0.50\linewidth]{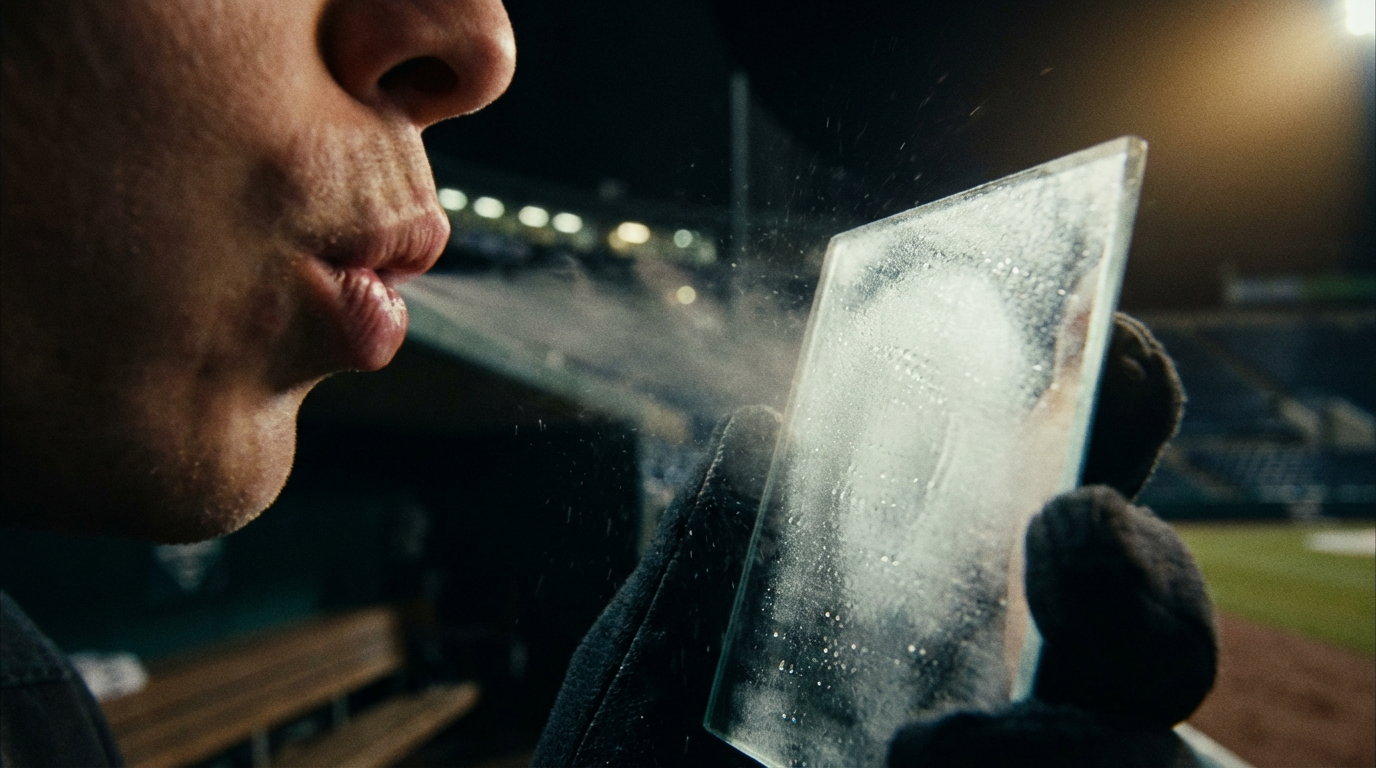}
  \caption{First-frame input used for I2V generation. \textbf{\textit{Prompt}}: ``The man exhales a warm breath onto the glass.''}
\end{figure}

\clearpage
\noindent\textbf{\color{blue}Task 33: Life \& Healthcare Knowledge}

\noindent\textbf{Definition:} Given a medical scene, the model is required to generate changes based on biological responses and basic medical logic.

\begin{figure}[htbp]
  \centering
  \includegraphics[width=0.50\linewidth]{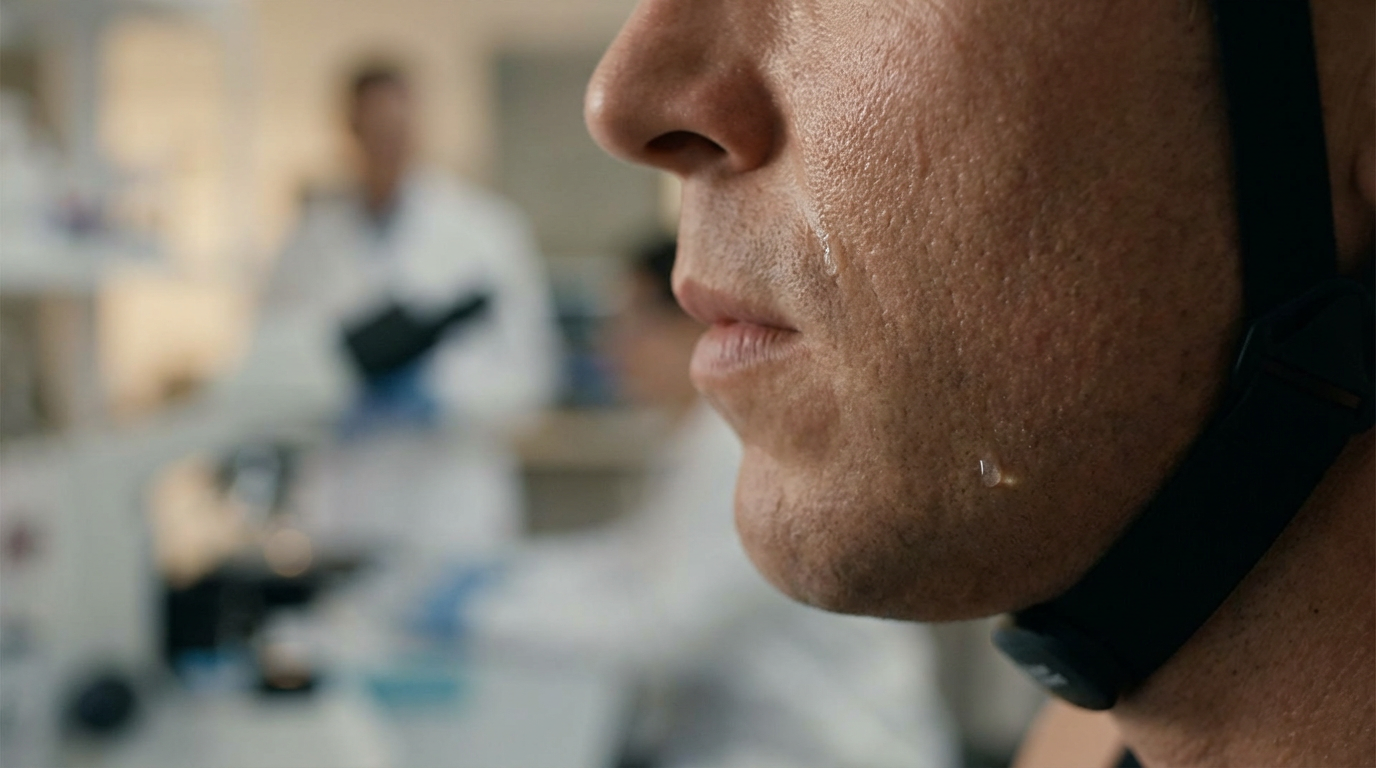}
  \caption{First-frame input used for I2V generation. \textbf{\textit{Prompt}}: ``The skin's reaction as the intense exercise continues.''}
\end{figure}

\noindent\textbf{\color{blue}Task 34: Geography \& Earth Science}

\noindent\textbf{Definition:} Given a scene, the model is required to generate changes involving macroscopic phenomena such as river formation, celestial systems, and weather evolution.

\begin{figure}[htbp]
  \centering
  \includegraphics[width=0.50\linewidth]{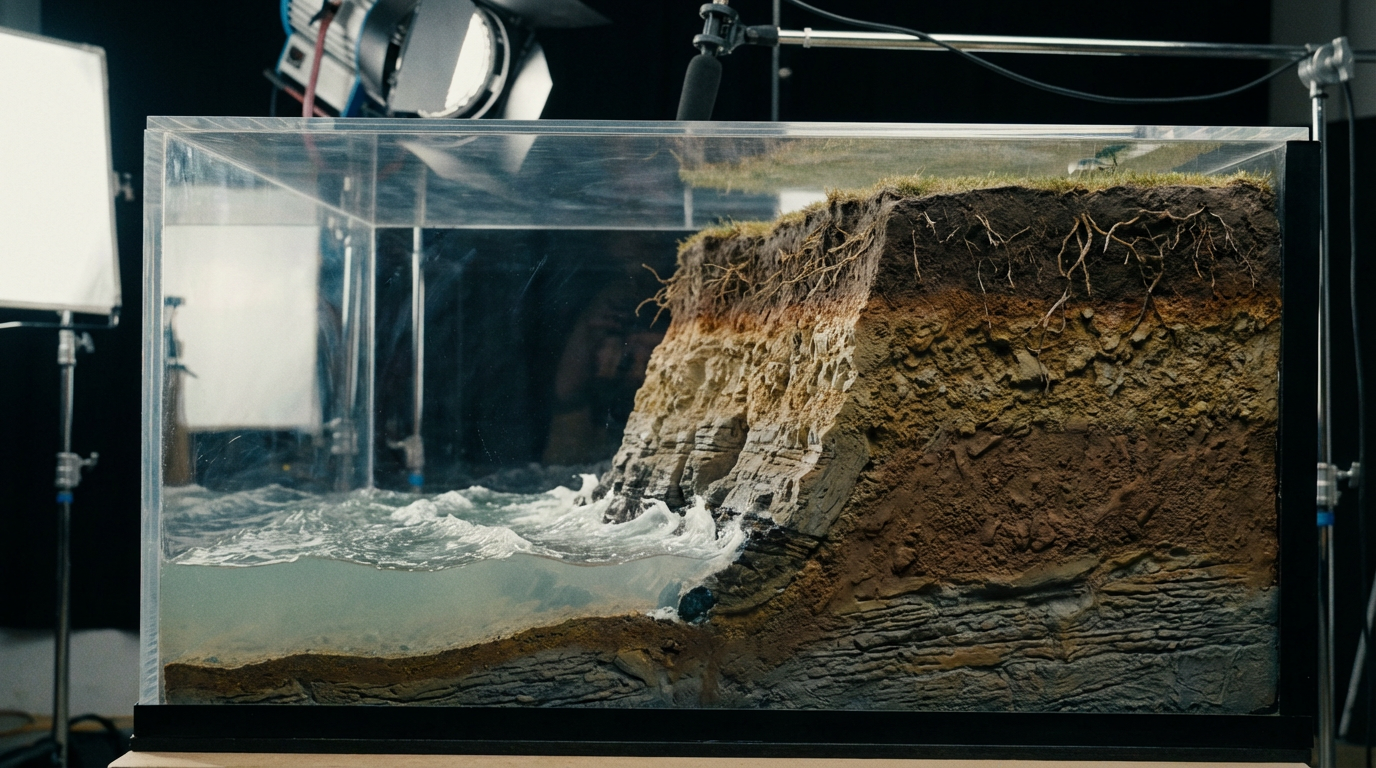}
  \caption{First-frame input used for I2V generation. \textbf{\textit{Prompt}}: `Showing tectonic plates pushing against each other.''}
\end{figure}

\clearpage
\noindent\textbf{\color{blue}Task 35: Specialized Sports Kinematics}

\noindent\textbf{Definition:} Given a sports scene and several people, the model is required to make them perform sports actions with professional characteristics.

\begin{figure}[htbp]
  \centering
  \includegraphics[width=0.50\linewidth]{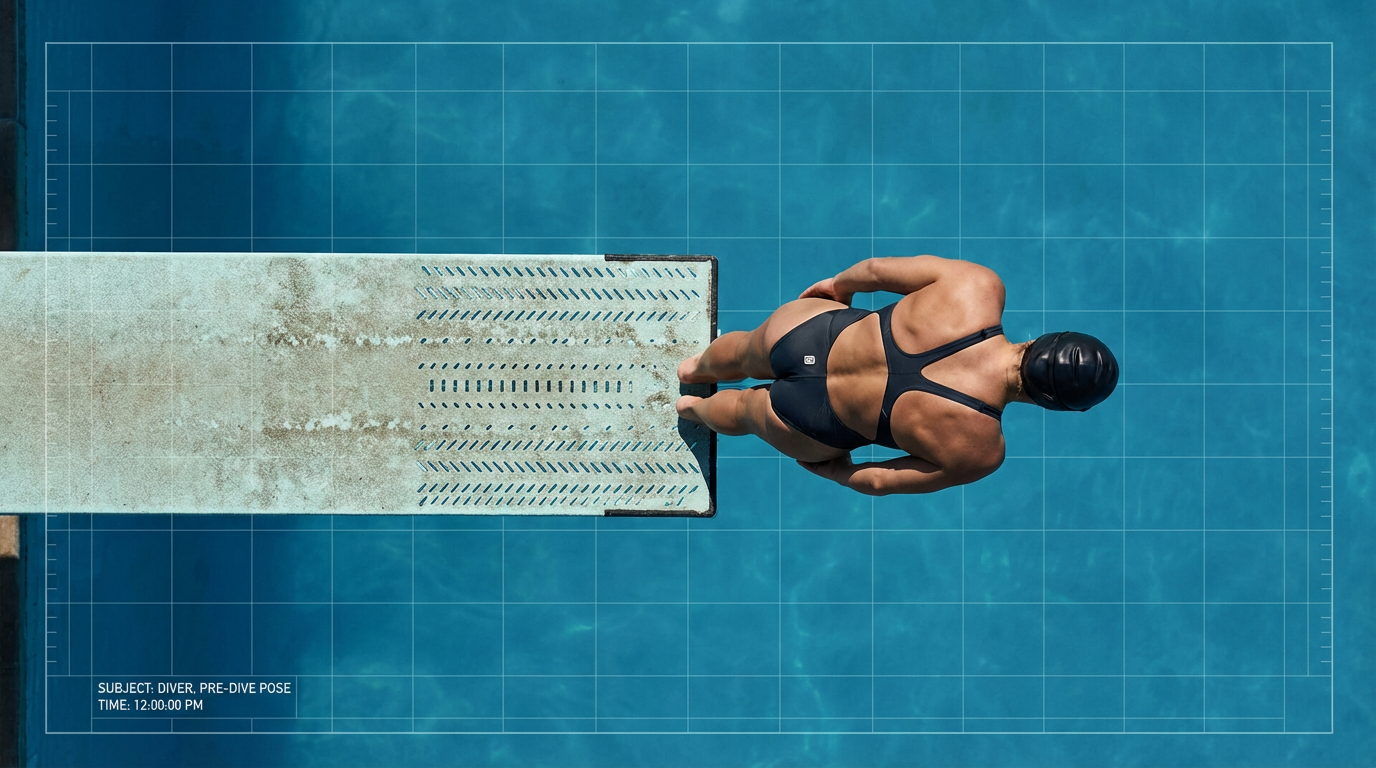}
  \caption{First-frame input used for I2V generation. \textbf{\textit{Prompt}}: ``The athlete jumps upward.''}
\end{figure}

\noindent\textbf{\color{blue}Task 36: Social Rules}

\noindent\textbf{Definition:} Given a social scene, the model is required to control the people in the scene or a first-person protagonist to carry out instructions grounded in social contracts, norms, and the implicit ``script'' of the specific scenario.

\begin{figure}[htbp]
  \centering
  \includegraphics[width=0.50\linewidth]{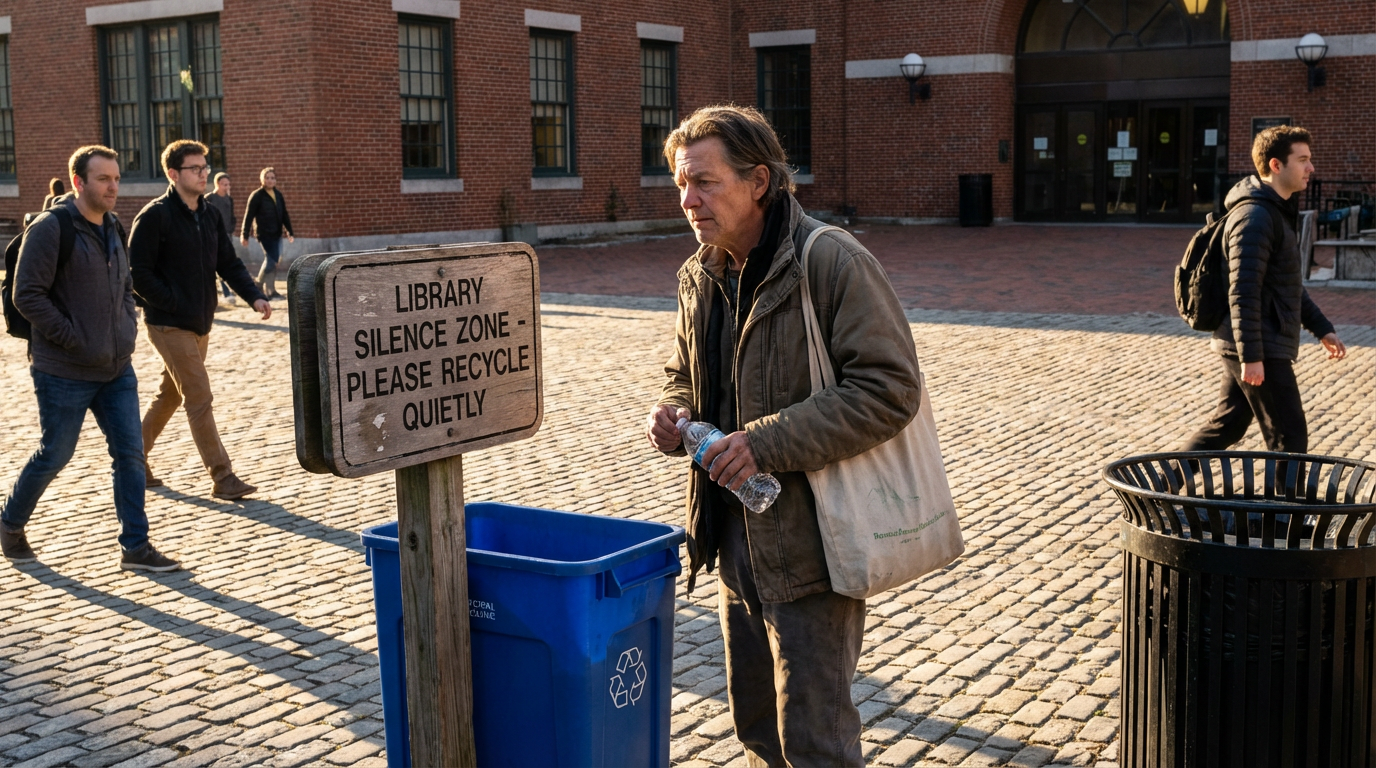}
  \caption{First-frame input used for I2V generation. \textbf{\textit{Prompt}}: ``The man should dispose of the empty bottle.''}
\end{figure}

\clearpage
\noindent\textbf{\color{blue}Task 37: Cultural Customs}

\noindent\textbf{Definition:} Given a real-world scene, the model is required to carry out instructions involving specific cultural symbols, festival customs, and metaphorical abstract concepts.

\begin{figure}[htbp]
  \centering
  \includegraphics[width=0.50\linewidth]{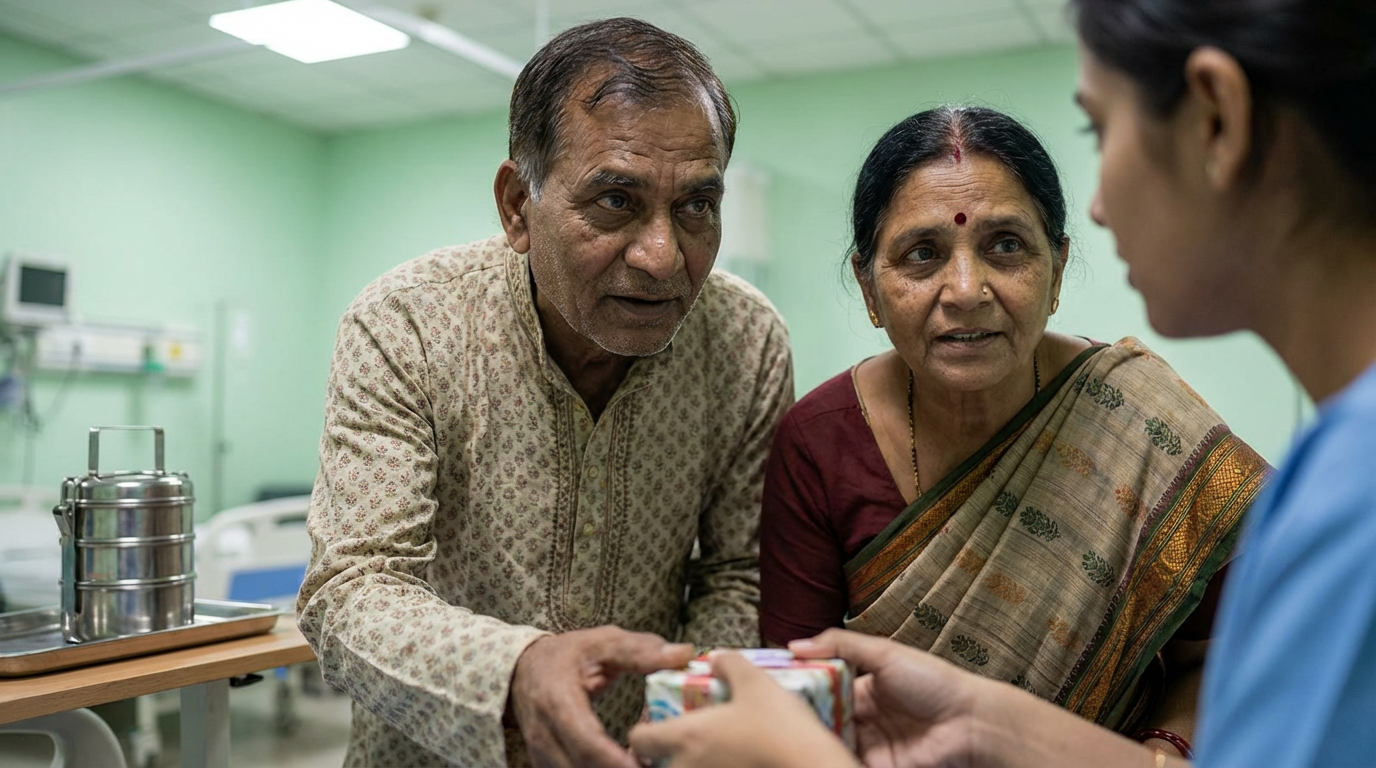}
  \caption{First-frame input used for I2V generation. \textbf{\textit{Prompt}}: ``The couple greets the visitor.''}
\end{figure}

\noindent\textbf{\color{blue}Task 38: Emotion Recognition}

\noindent\textbf{Definition:} Given a person, the model is required to make them perform behaviors involving emotional changes, micro-expressions, and body language.

\begin{figure}[htbp]
  \centering
  \includegraphics[width=0.50\linewidth]{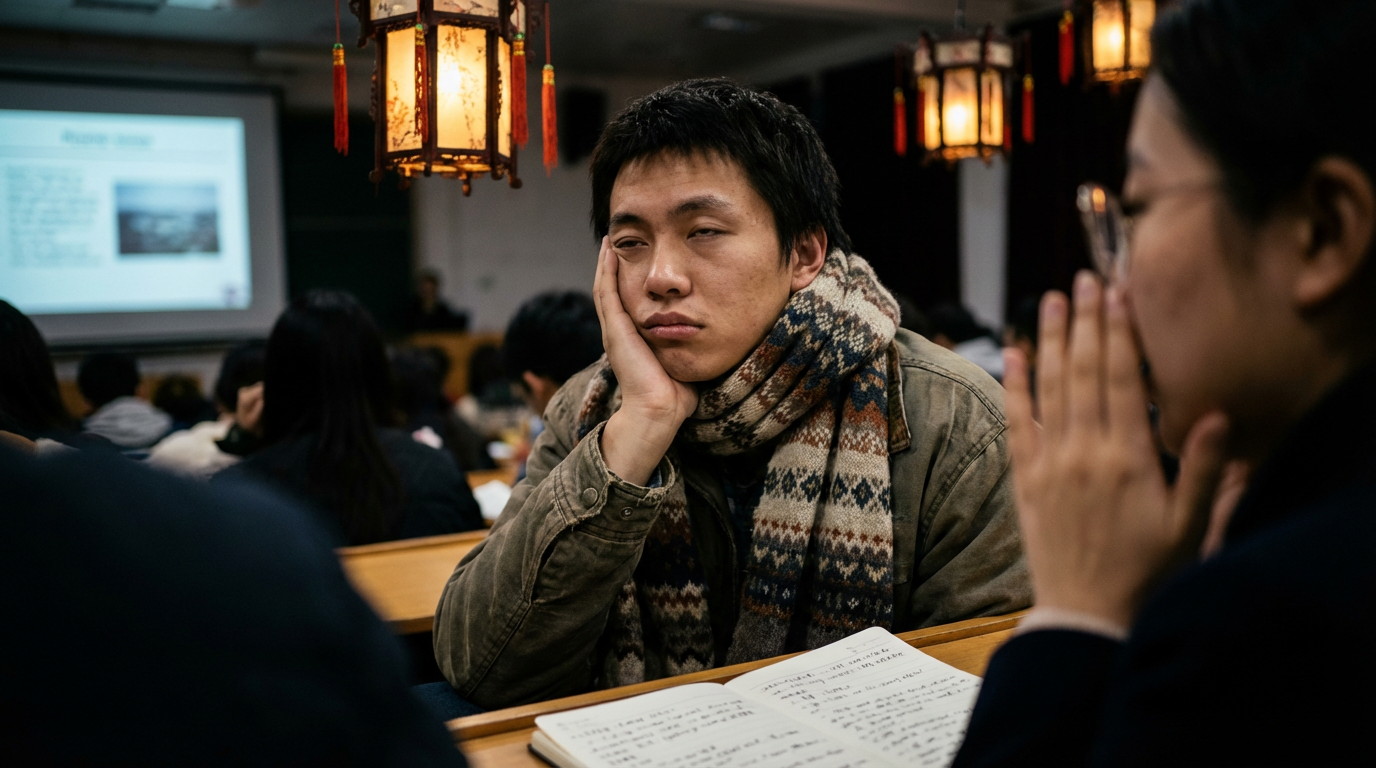}
  \caption{First-frame input used for I2V generation. \textbf{\textit{Prompt}}: ``Showing their natural reaction as the dull lecture drags on.''}
\end{figure}

\clearpage

\section{VWG-Bench Prompt Templates}

\begin{promptbox}{Last Frame Evaluation}
You are a professional engineer in the field of image-to-video generation.
You will be given the last frame of a video generated by an image-to-video model for the \{task\_name\} task, along with a pre-designed ``Last frame goal'' that describes what should appear in the last frame.
Your task is to evaluate how consistent this last frame is with the given ``Last frame goal'', assign a score from 1 to 5 (from least consistent to most consistent), and provide the reason for the score you assign.

The ``Last frame goal'': \{last\_frame\_goal\}

Output your result in the following format:

\texttt{\{"last\_frame\_goal\_score": "<your score here>", "reason\_for\_last\_frame\_goal\_score": "<your reason here>"\}}
\end{promptbox}

\begin{promptbox}{Video Evaluation}
You are a professional engineer in the field of image-to-video generation.
You will be given a video generated by an image-to-video model for the \{task\_name\} task with the image-to-video prompt: ``\{user\_prompt\}''.
Your task is to evaluate the following metrics based on the corresponding scoring criteria: \texttt{video\_quality}, \texttt{progress\_consistency}, (Optional) \texttt{implicit\_rule}, and (Optional) \texttt{progress\_goal}.

\textbf{First}, evaluate the overall quality of the video, focusing only on the visual and temporal coherence of the frames themselves (not semantics, prompt adherence, or aesthetics). Assess stability and continuity, including but not limited to: abrupt visual shifts, jump cuts, frame drops or repeats, jitter or stutter, flicker (brightness, color, or texture), temporal inconsistency (identity, shape, or size drifting), warping or melting, ghosting or double edges, motion discontinuities, camera instability, rolling-shutter-like distortions, aliasing or shimmering, compression or blocking artifacts, noise bursts, sudden blur changes, and background ``breathing'' or deformation. Penalize any brief but severe failures strongly. Assign a score from 1 to 5 (1 = unusable or very unstable with frequent severe artifacts; 2 = major artifacts that substantially disrupt viewing; 3 = noticeable artifacts but generally watchable; 4 = minor occasional artifacts with mostly stable frames; 5 = very stable, clean, and temporally coherent), and provide the reason for the score you assign.

\textbf{Second}, evaluate the overall temporal consistency of the video, that is, whether objects or scenes in the video undergo changes that should not occur. The specific rule is ``\{foreground\}'' (Optional), ``\{background\}''. Based on how consistent the video is with this rule, assign a score from 1 to 5 (from least consistent to most consistent), and provide the reason for the score you assign.

\textbf{(Optional) Third}, evaluate how well the video understands the task, that is, whether the video follows the implicit rules of the task. The implicit rule of this task is ``\{implicit\_rule\}''. Based on how consistent the video is with this implicit rule, assign a score from 1 to 5 (from least consistent to most consistent), and provide the reason for the score you assign.

\textbf{(Optional) Finally}, evaluate whether the process shown in the video meets expectations, that is, whether the video aligns with the predefined ``Progress goal''. This ``Progress goal'' is: ``\{progress\_goal\}''. Based on the consistency between the video and the ``Progress goal'', assign a score from 1 to 5 (from least consistent to most consistent), and provide the reason for the score you assign.

Output your evaluation result in the following format:

\texttt{\{}

\texttt{\quad "video\_quality\_score": "<your video\_quality score here>",}

\texttt{\quad "reason\_for\_video\_quality\_score": "<your reason here>",}

\texttt{\quad "progress\_consistency\_score": "<your progress\_consistency score here>",}

\texttt{\quad "reason\_for\_progress\_consistency\_score": "<your reason here>",}

\texttt{\quad (Optional) "implicit\_rule\_score": "<your implicit\_rule score here>",}

\texttt{\quad (Optional) "reason\_for\_implicit\_rule\_score": "<your reason here>",}

\texttt{\quad (Optional) "progress\_goal\_score": "<your progress\_goal score here>",}

\texttt{\quad (Optional) "reason\_for\_progress\_goal\_score": "<your reason here>"}

\texttt{\}}
\end{promptbox}

\begin{promptbox}{Last Frame Goal Annotation}
You will be given a prompt for an image-to-video task and its corresponding first-frame image.
Your task is to obtain a ``last frame goal'', that is, by imagining or reasoning, determine what the final frame of the video generated from the given first-frame image and image-to-video prompt should look like (for example, if the first frame shows the interior of a house that is dimly lit with a light switch in the center of the image, and the image-to-video prompt is to turn on the light, then the ``last frame goal'' should be ``the switch is turned on and the interior becomes bright'').
The format should be concise, in the form of a single sentence.

The image-to-video prompt: \{user\_prompt\}

Output your ``last frame goal'' in the following format:

\texttt{\{"last\_frame\_goal": "<your last frame goal here>"\}}
\end{promptbox}

\begin{promptbox}{Text-to-Image Output Filtering}
Please determine whether the prompt in this text-to-image output perfectly matches the corresponding generated image.
That is, all elements appearing in the prompt must be present in the image, and no elements not mentioned in the prompt should appear in the image.
If there is no perfect match, set \texttt{result} to 0 and provide a reason; otherwise, set \texttt{result} to 1 and set the reason to null.

The text-to-image prompt: \{t2i\_prompt\}

Output your result and reason in the following format:

\texttt{\{"result": "<your result here>", "reason": "<your reason here>" or null\}}
\end{promptbox}

\clearpage
\section{Vid-PRE Prompt Templates}

\begin{promptbox}{Chain-of-Thought and Optimized Prompt Annotation}
You are a professional image-to-video prompt engineer.
Your task is to produce a chain-of-thought and an optimized prompt, given ONE static image (first frame) and ONE original prompt for an image-to-video model.

Write the chain-of-thought in a natural, human style: free-form prose, varied sentence lengths, no numbered lists or headings, and no rigid section titles. It should read like someone genuinely thinking through the problem, so light hesitations such as ``hmm'', ``oh'', ``well'', and ``okay, so'' are allowed as long as they do not dominate the text. Do not follow a fixed template; let the narrative flow adapt to each case.

In the chain-of-thought, casually but clearly cover: what you infer from the image at a high level; how you reinterpret and clarify the user's intent; any minimal assumptions you decide to make; the key rules and constraints that should govern motion, camera or viewport behavior, and what must stay unchanged; when the sequence should end; and what counts as success versus failure for this image-to-video generation. You may be detailed and reflective, but keep the chain-of-thought within about 384 tokens. Avoid bullet points and explicit enumeration such as ``1), 2), 3)''.

Then write the optimized prompt as a single, self-contained instruction for a commercial image-to-video model: imperative, unambiguous, and concise compared to the chain-of-thought, while still fully specifying the needed behavior. Explicitly encode the semantics you rely on from the image, motion or behavior rules, camera behavior, invariants, and the end condition. Avoid meta-commentary, filler words, and examples in the optimized prompt.

General rules: do not request extra inputs; if you assume something, state it in the chain-of-thought and reflect it succinctly in the optimized prompt; do not invent new visual assets that contradict the image or user intent; prefer operational, checkable wording over purely subjective adjectives. Avoid using unescaped double quotes inside the JSON strings.

Output strictly in JSON with two fields.

Original prompt: \{original\_prompt\}

Output your chain-of-thought and optimized prompt in the following format:

\texttt{\{"chain\_of\_thought": "<your chain-of-thought here>", "optimized\_prompt": "<your optimized prompt here>"\}}
\end{promptbox}

\begin{promptbox}{Rejection Sampling}
This is a task to determine whether an image-to-video prompt is appropriate and valid.
You will be given an image as the first frame of this image-to-video task, a text describing the task objective (which can be considered the ground truth), and an annotated image-to-video prompt.
You need to examine the image-to-video prompt for: whether all content mentioned in the ground truth text is implemented in the image-to-video prompt (i.e., whether it is consistent with the ground truth text); and whether the spatial relationships mentioned in the image-to-video prompt, such as ``left'' and ``bottom right'', are consistent with the spatial relationships in the image (i.e., consistency with the first frame).
Finally, please give the image-to-video prompt a comprehensive score on both aspects of consistency, using a scale from 1 to 5 (from extremely inappropriate to extremely appropriate), and then provide a reason to support your score.

The ground truth text: \{gt\_text\}

The image-to-video prompt: \{optimized\_prompt\}

Output your score in the following format:

\texttt{\{"score": "<your score here>", "score\_reason": "<your reason here>"\}}
\end{promptbox}

\begin{promptbox}{Rule-Based Final Prompt Optimization}
This is a task to optimize an image-to-video prompt based on given rules.
You will be given an image-to-video prompt and a text containing a set of rules designed to improve the prompt's output.
You need to append these rule texts naturally to the image-to-video prompt with minimal modifications, adding the final ``patch'' to the prompt.

The rule text: \{rule\}

The image-to-video prompt: \{optimized\_prompt\}

Output your optimized prompt in the following format:

\texttt{\{"optimized\_prompt": "<your optimized image-to-video prompt here>"\}}
\end{promptbox}

\begin{promptbox}{Reward Scoring for Reinforcement Learning}
You are a strict-but-pragmatic evaluator for image-to-video prompt rewriting.
You will be given a first-frame image, an original image-to-video prompt, and a rewritten prompt.
The rewritten prompt should be an expansion or refinement of the original prompt.

\textbf{EVALUATION PHILOSOPHY (VERY IMPORTANT):}

A) Prioritize CONTENT requirements over CAMERA terminology.

B) Be tolerant to camera wording when the original prompt is ambiguous or contradictory.

C) The most important improvement is adding TASK-CRITICAL DETAILS (decision-making details), not generic quality constraints.

\textbf{Step 1:} Extract the ORIGINAL prompt into two lists:

1) NON-NEGOTIABLE CONTENT REQUIREMENTS (highest priority): key objects or regions to target, required edits or overlays (arrows, boxes, glow, text), and any measurement or fixed-scale constraints.

2) SECONDARY CAMERA REQUIREMENTS (lower priority): static, zoom, pan, or dolly wording.

\textbf{Step 2:} Check coverage in the rewritten prompt:
\begin{itemize}
    \item For each NON-NEGOTIABLE CONTENT requirement, mark: preserved, partially preserved, missing, or substituted.
    \item For camera requirements, mark: consistent, acceptable reinterpretation, or conflicting.
\end{itemize}

\textbf{Step 3:} Identify NEW DETAILS added by the rewritten prompt and classify each as either:
\begin{itemize}
    \item TASK-CRITICAL detail (high value): concrete, verifiable decisions that directly determine successful completion (e.g., exact moves or paths, which cell gets which number, which puzzle piece goes to which slot or region, exact arrow or box placement).
    \item SUPPORTING detail (medium or low value): generic quality constraints or anti-failure constraints (e.g., keep background fixed, only object X moves, smooth motion, preserve lighting, maintain geometry).
\end{itemize}

\textbf{SCORING RULES:}

1) Intent preservation (1--5): Score mainly based on whether NON-NEGOTIABLE CONTENT requirements are preserved. If ANY key content requirement is missing or substituted, the intent preservation score should be $\leq 3$. Do NOT strongly penalize minor camera wording differences when the content goal is preserved.

2) Image consistency (1--5): Determine whether the mentioned objects or regions are actually visible and correctly referenced. Penalize hallucinations or incorrect spatial relations.

3) Solution correctness (1--5): focus on TASK-CRITICAL NEW DETAILS and whether they are correct. Use this rubric:
\begin{itemize}
    \item 5: Adds MULTIPLE correct TASK-CRITICAL details that substantially reduce ambiguity and make the task almost directly executable.
    \item 4: Adds at least ONE correct TASK-CRITICAL detail that clearly improves executability, with no major harmful errors.
    \item 3: Neutral or modest refinement: adds only SUPPORTING details, or adds correct but non-critical details that do not materially change executability.
    \item 2: Adds incorrect or conflicting details (including incorrect task-critical details) that would likely reduce success.
    \item 1: The added details are largely wrong or make the task fail.
\end{itemize}

\textbf{IMPORTANT coupling rule:} If \texttt{intent\_preservation\_score} $\leq 3$ because key content requirements are missing or substituted, then \texttt{solution\_correctness\_score} MUST also be $\leq 3$.

The original image-to-video prompt: \{origin\_user\_prompt\}

The rewritten prompt: \{user\_prompt\}

Output JSON in the following format (strings must be properly escaped):

\texttt{\{}

\texttt{\quad "intent\_preservaion\_score": <1-5>,}

\texttt{\quad "reason\_for\_intent\_preservaion\_score": "<cite Step 2 coverage>",}

\texttt{\quad "image\_consistency\_score": <1-5>,}

\texttt{\quad "reason\_for\_image\_consistency\_score": "<reason>",}

\texttt{\quad "solution\_correctness\_score": <1-5>,}

\texttt{\quad "reason\_for\_solution\_correctness\_score": "<whether NEW details are task-critical vs supporting, and whether they are correct>"}

\texttt{\}}
\end{promptbox}

\fi
\end{document}

%% file: sections/1_introduction.tex
% ----------------------------------------------------------------
%  SECTION 1 — INTRODUCTION
% ----------------------------------------------------------------

\section{Introduction}
\label{sec:intro}

% ── Paragraph 1: The paradigm shift and the central question ──────────────────
Reasoning through continuous spatio-temporal signals is a hallmark of
intelligence that text and static images cannot fully replicate~\cite{lecun2022path, zhou2025vlm4d, ding2025understanding,luo2026dr}.
``Thinking with Text'' established Chain-of-Thought (CoT)~\cite{wei2022chain, lyu2023faithful, zhang2024chain,luo2026unveiling}
prompting as a foundational paradigm for language reasoning; ``Thinking with
Images'' extended this to visual chains in Vision-Language
Models~\cite{zheng2025deepeyes, su2025thinking, li2025tir}.
A natural next frontier is \emph{Think with Video}: reasoning expressed through
continuous video frames that encode physical dynamics, object interactions, and
long-horizon goal execution in a unified temporal substrate.
Recent advances, including Veo~3's chain-of-frames, Sora-2's visual problem
solving~\cite{tong2025thinking, ghazanfari2025chain, liu2025can}, and emergent spatial planning in open-source
models, suggest that video generation is crossing a
threshold from photorealistic rendering toward genuine world simulation~\cite{cho2024sora, lin2025exploring, wang2025survey}.
Yet a fundamental question remains unresolved: do these models truly
understand the rules they appear to follow, or are they exploiting
surface statistics to produce plausible-looking sequences?

% ── Paragraph 2: The evaluation gap ─────────────────────────────────────────
Answering this question requires benchmarks that probe reasoning rather than
aesthetics.
Existing efforts have made important strides:
TiViBench~\cite{chen2025tivibench} introduces hierarchical evaluation across
four structural and logical dimensions;
V-ReasonBench~\cite{luo2025v} provides reproducible, answer-verifiable tasks
grounded in spatial and physical cognition;
VR-Bench~\cite{yang2025reasoning} systematically measures spatial planning
through maze-solving;
RULER-Bench~\cite{he2025ruler} targets cognitive-rule adherence across six
categories;
RISE-Video~\cite{liu2026rise} probes implicit world-rule understanding
across eight knowledge domains; and
MMGR~\cite{cai2025mmgr} covers abstract and embodied reasoning.
Despite this progress, the evaluation landscape remains fragmented.
Each benchmark covers only a subset of the reasoning spectrum.
Moreover, existing protocols typically collapse evaluation into a single
binary or holistic score, masking where and why models fail, specifically whether at the level of visual fluency, task-level rule tracking, or precise goal realization.

\begin{figure*}[!t]
    \centering
    \includegraphics[width=0.90\linewidth]{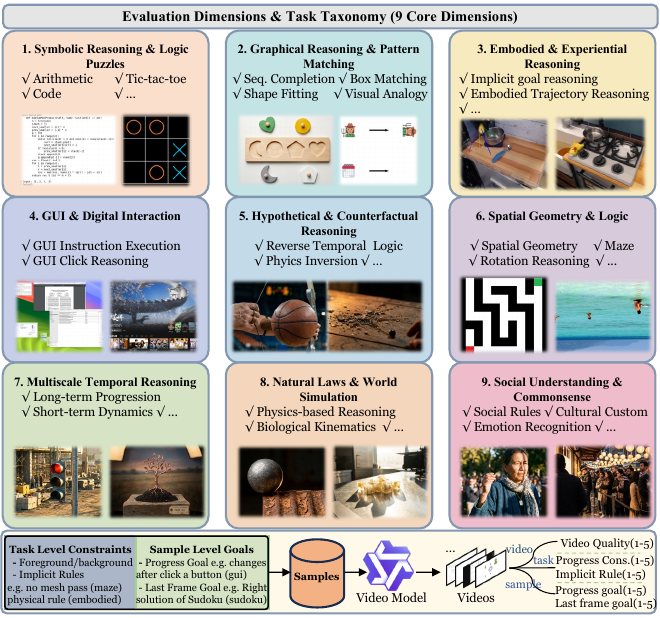}
    \caption{Comprehensive overview of \texttt{VWG-Bench}'s taxonomy, automated data construction, and structured evaluation framework.}
    \label{fig:teaser}
\end{figure*}

% ── Paragraph 3: Our benchmark ───────────────────────────────────────────────
To close both gaps, we introduce \textbf{\texttt{VWG-Bench}}, a comprehensive and hierarchical benchmark designed to
stress-test think-with-video reasoning across the full spectrum of world
simulation.
As illustrated in Figure~\ref{fig:teaser}, \texttt{VWG-Bench} covers
\textbf{9} core reasoning dimensions and \textbf{38} fine-grained tasks.
Specifically, these dimensions systematically evaluate:
% (1)~\textbf{Symbolic Reasoning \& Logic Puzzles}, testing the strict execution
% of algorithmic rules and arithmetic calculations;
% (2)~\textbf{Graphical Reasoning \& Pattern Matching}, requiring models to
% identify visual rules and complete spatio-temporal sequences;
% (3)~\textbf{Embodied \& Experiential Reasoning}, assessing intention causality
% and precise execution of complex procedural trajectories;
% (4)~\textbf{GUI \& Digital Interaction}, simulating correct state transitions
% in digital interfaces driven by user commands;
% (5)~\textbf{Hypothetical \& Counterfactual Reasoning}, challenging models to
% perform zero-shot logical deduction under rewritten rules or reversed time;
% (6)~\textbf{Spatial Geometry \& Logic}, verifying 3D spatial transformations
% and topological consistency;
% (7)~\textbf{Multiscale Temporal Reasoning}, probing state evolution from
% short-term dynamics to long-horizon processes;
% (8)~\textbf{Natural Laws \& World Simulation}, testing the internalization of
% objective laws across physics, biology, and specialized kinematics; and
% (9)~\textbf{Social Understanding \& Commonsense}, comprehending human-centric
% scripts, emotional causality, and cultural norms.
(1)~Symbolic Reasoning \& Logic Puzzles,
(2)~Graphical Reasoning \& Pattern Matching,
(3)~Embodied \& Experiential Reasoning,
(4)~GUI \& Digital Interaction,
(5)~Hypothetical \& Counterfactual Reasoning,
(6)~Spatial Geometry \& Logic,
(7)~Multiscale Temporal Reasoning,
(8)~Natural Laws \& World Simulation, and
(9)~Social Understanding \& Commonsense.
To construct evaluation samples at scale, we develop a highly automated data pipeline that orchestrates image generation, consistency
filtering, and intent formulation,
enabling scalable production of high-quality instances.

% ── Paragraph 4: Our evaluation protocol ─────────────────────────────────────
Beyond scale and coverage, \texttt{VWG-Bench} introduces a
three-level VLM-as-Judge evaluation framework with a fine-grained
1--5 scoring scale.
Unlike existing holistic scoring that conflates rendering quality with
cognitive correctness, our protocol explicitly decouples three orthogonal
facets of performance:
\emph{Video-level}, assessing overall visual fluency and temporal coherence;
\emph{Task-level}, measuring progress consistency and adherence to implicit
task rules across frames; and
\emph{Sample-level}, evaluating the precise realization of intermediate
milestones and final objectives.
This decomposition enables fine-grained, interpretable diagnostics that
identify failure modes at each level of the reasoning hierarchy.
Extensive evaluations of leading commercial and open-source models on
VWG-Bench expose a consistent pattern: models achieve strong
Video-level rendering scores but suffer severe degradation at the Task- and
Sample-level on compositional, rule-constrained, and logic-heavy scenarios.
This gap is not primarily a failure of visual generation; it is a failure of
reasoning through prompts: given only a natural-language description,
models lack the structured representation of task rules, success criteria, and
constraint hierarchies needed to guide generation toward correct reasoning
trajectories.

% ── Paragraph 6: Vid-PRE ─────────────────────────────────────────────────────
To directly address this limitation, we propose \textbf{\texttt{Vid-PRE}}, a model-agnostic prompt rewriter that
decouples the reasoning stage from the rendering stage.
\texttt{Vid-PRE} offloads the cognitive burden to a dedicated VLM, which
infers task rules and success criteria from the image context and distills
them into a concise, constraint-aware prompt for the downstream video
generator.
\texttt{Vid-PRE} is trained via Supervised Fine-tuning (SFT) on rejection-sampled reasoning tuples and
further optimized with Group Relative Policy Optimization (GRPO)~\cite{shao2024deepseekmath} using a meticulously designed purely text-based rewards system, keeping it agnostic to any specific generator architecture.
Extensive experiments confirm substantial improvements across multiple
generators without any architectural modification.

% ── Paragraph 7: Contributions ───────────────────────────────────────────────
In summary, our contributions are threefold:
\begin{itemize}[leftmargin=*]
    \item \textbf{A Unified Think-with-Video Benchmark.}
    \texttt{VWG-Bench} is the first benchmark to jointly cover all nine core
    dimensions of think-with-video world simulation across 38 tasks, constructed
    via a fully automated data pipeline for scalable and reproducible
    scenario generation.

    \item \textbf{A Fine-Grained, Hierarchical Evaluation Protocol.}
    Our three-level VLM-as-Judge framework with a 1--5 scoring scale
    diagnoses Video-level fluency, Task-level rule adherence,
    and Sample-level goal realization, revealing failure modes that holistic
    or binary evaluation cannot.

    \item \textbf{A Universal and Plug-and-Play Prompt Reasoner and Enhancer.}
    \texttt{Vid-PRE} is a model-agnostic rewriter trained via SFT and GRPO
    with text-based rewards that significantly boosts reasoning accuracy across
    multiple video generators, without any modification to their architectures.
\end{itemize}

%% file: sections/2_benchmark.tex
\section{\texttt{VWG-Bench}: Video World Generalist Benchmark}
\label{sec:benchmark}
\subsection{Design Principles}

VWG-Bench is designed to comprehensively evaluate the reasoning capabilities of video generation models. Compared with prior reasoning benchmarks for video generation, VWG-Bench aims to support both broader task coverage and a more systematic evaluation framework. 

For task coverage, as shown in Table~\ref{tab:benchmark_comparison}, our benchmark spans nine major reasoning dimensions and 38 task categories, fully covering symbolic and graphical reasoning, embodied and interactive tasks, as well as spatial--temporal reasoning, physical world simulation, and social commonsense understanding. To balance evaluation comprehensiveness with practical cost, we sample only 10 high-quality instances from each task category, significantly reducing the computational and annotation overhead required for evaluation.

For the evaluation framework, as illustrated in Figure~\ref{fig:teaser}, we establish a three-level evaluation protocol. The video-level performs coarse-grained quality control to ensure the generated video is a valid visual medium. The task-level verifies whether the model respects task-specific constraints and logical rules throughout the video generation process. Finally, the sample-level assesses whether the final outcome satisfies the intended task objective. Compared to prior benchmarks that typically involve only one or two evaluation dimensions, our framework provides a more comprehensive and structured assessment of reasoning performance.

\subsection{Evaluation Dimensions and Task Taxonomy}
\label{subsec:taxonomy}

Here we introduce the nine evaluation dimensions and the 38 task categories covered by VWG-Bench.

VWG-Bench organizes tasks into nine reasoning dimensions spanning symbolic logic, visual pattern induction, embodied interaction, digital interfaces, counterfactual reasoning, spatial geometry, temporal evolution, world simulation, and social understanding.
\textbf{Symbolic Reasoning \& Logic Puzzles} evaluate the execution of formal rules and algorithms, such as ``Arithmetic'', ``Coding'', and ``Sudoku''.
\textbf{Graphical Reasoning \& Pattern Matching} focuses on discovering visual correspondences and completing structured patterns, such as ``Box Matching'', ``Color Connecting'', and ``Visual Analogy''.
\textbf{Embodied \& Experiential Reasoning} examines reasoning in physical interaction scenarios, including tasks like ``Embodied Action Execution'', ``Embodied Trajectory Reasoning'', and ``Implicit Goal Reasoning''.
\textbf{GUI \& Digital Interaction} evaluates reasoning within graphical user interfaces, such as ``GUI Instruction Execution'' and ``GUI Click Reasoning''.
\textbf{Hypothetical \& Counterfactual Reasoning} probes reasoning under altered or fictional premises, for example ``Reverse Temporal Logic'', ``Physics Inversion'', and ``Causal Intervention''.
\textbf{Spatial Geometry \& Logic} tests geometric consistency and spatial transformations, including tasks such as ``Maze'', ``Rotation Reasoning'', and ``Spatial Geometry''.
\textbf{Multiscale Temporal Reasoning} evaluates the ability to maintain coherent dynamics across different time horizons, such as ``Short-term Temporal Dynamics'' and ``Long-term Temporal Progression''.
\textbf{Natural Laws \& World Simulation} measures whether models follow real-world physical or biological principles, for example ``Physics-based Reasoning'', ``Biological Kinematics'', and ``Physical Commonsense Reasoning''.
\textbf{Social Understanding \& Commonsense} focuses on human-centric semantics and social behavior modeling, including tasks like ``Social Rules'', ``Cultural Customs'', and ``Emotion Recognition''.

% ================================================================
%  VWG-Bench vs. Related Benchmarks — Top-Conference Style Table
% ================================================================

\newcommand{\cmark}{\textcolor[RGB]{34,139,34}{\ding{51}}}
\newcommand{\xmark}{\textcolor[RGB]{180,0,0}{\ding{55}}}
\newcommand{\half}{\textcolor[RGB]{200,140,0}{$\circ$}}
\definecolor{ourrow}{RGB}{232,240,255}
\definecolor{headergray}{RGB}{240,240,240}

% ----------------------------------------------------------------
% ----------------------------------------------------------------
\begin{table*}[t]
\centering
\caption{
    Comparison of \texttt{VWG-Bench} with representative video-generation reasoning benchmarks.
    \textbf{Sym}: Symbolic \& Logic,
     \textbf{Gph}: Graphical \& Pattern,
     \textbf{Emb}: Embodied \& Experiential,
     \textbf{GUI}: GUI \& Digital Interaction,
     \textbf{Cft}: Hypothetical \& Counterfactual,
     \textbf{Spa}: Spatial Geometry,
     \textbf{Tmp}: Multiscale Temporal,
     \textbf{Phy}: Natural Laws \& Physics,
     \textbf{Soc}: Social \& Commonsense).
    \textbf{Auto} = whether an automated data-construction pipeline is provided.
    \cmark~= fully covered; \half~= partially covered; \xmark~= not covered.
}
\label{tab:benchmark_comparison}
\setlength{\tabcolsep}{5pt}
\renewcommand{\arraystretch}{1.30}
\resizebox{\textwidth}{!}{%
\begin{tabular}{l ccc ccccccccc cc}
\toprule

% ══ Header row 1: group labels ════════════════════════════════════
& \multicolumn{3}{c}{\cellcolor{headergray}\textbf{Scale}}
& \multicolumn{9}{c}{\cellcolor{headergray}\textbf{Reasoning Coverage}}
& \multicolumn{2}{c}{\cellcolor{headergray}\textbf{Evaluation}}
\\
\cmidrule(lr){2-4}
\cmidrule(lr){5-13}
\cmidrule(lr){14-15}

% ══ Header row 2: column names ════════════════════════════════════
\textbf{Benchmark}
  & \textbf{Dims} & \textbf{Tasks} & \textbf{Samples}
  & \textbf{Sym} & \textbf{Gph} & \textbf{Emb} & \textbf{GUI}
  & \textbf{Cft} & \textbf{Spa} & \textbf{Tmp} & \textbf{Phy} & \textbf{Soc}
  & \textbf{Levels} & \textbf{Auto}
\\
\midrule

MME-CoF~\cite{guo2025video}
  & / & 12 & 59
  & \xmark & \xmark & \half & \cmark
  & \xmark & \cmark & \half & \cmark & \xmark
  & 1  & \xmark \\

VideoThinkBench~\cite{tong2025thinking}
  & 2 & 5 & 4{,}149
  & \half & \cmark & \xmark & \xmark
  & \xmark & \half & \xmark & \half & \xmark
  & 1  & \xmark \\

TiViBench~\cite{chen2025tivibench}
  & 4 & 24 & 595
  & \cmark & \cmark & \half & \xmark
  & \xmark & \cmark & \xmark & \xmark & \xmark
  & 1  & \xmark \\

VR-Bench~\cite{yang2025reasoning}
  & 1 & 5 & 7{,}920
  & \xmark & \xmark & \xmark & \xmark
  & \xmark & \cmark & \xmark & \xmark & \xmark
  & 1  & \xmark \\

V-ReasonBench~\cite{luo2025v}
  & 4 & 12 & 327
  & \cmark & \cmark & \xmark & \xmark
  & \xmark & \cmark & \xmark & \cmark & \xmark
  & 1  & \xmark \\

RULER-Bench~\cite{he2025ruler}
  & 6 & 40 & 622
  & \cmark & \xmark & \xmark & \xmark
  & \xmark & \cmark & \half & \cmark & \xmark
  & 1  & \xmark \\

MMGR~\cite{cai2025mmgr}
  & 3 & 10 & 1{,}853
  & \cmark & \xmark & \cmark & \xmark
  & \xmark & \cmark & \xmark & \cmark & \xmark
  & 1  & \xmark \\

RISE-Video~\cite{liu2026rise}
  & 8 & 29 & 467
  & \cmark & \half & \cmark & \xmark
  & \xmark & \cmark & \cmark & \half & \cmark
  & 1  & \cmark \\

\midrule

\rowcolor{ourrow}
\textbf{VWG-Bench (Ours)}
  & \textbf{9} & \textbf{38} & \textbf{380}
  & \cmark & \cmark & \cmark & \cmark
  & \cmark & \cmark & \cmark & \cmark & \cmark
  & \textbf{3} & \cmark \\

\bottomrule
\end{tabular}%
}
\end{table*}

\subsection{Data Construction Pipeline}
\label{subsec:data_construction}

\textbf{Image-Pair Initialization.}
To collect prompt-image pairs across such a diverse set of task categories, we adopt three complementary strategies: (1) identifying suitable open-source datasets and sampling high-quality categories from them (e.g., GUI and embodied tasks); (2) generating samples through programmatic algorithms (e.g., arithmetic and Sudoku tasks); or (3) automatically generating images via Text-to-Image (T2I) models using predefined prompt templates and dynamic variable pools to ensure broad scenario diversity. To guarantee visual integrity, we apply a strict visual filtering mechanism to T2I outputs. A VLM acts as an impartial judge to verify whether the generated image faithfully adheres to the T2I prompt without missing elements or visual hallucinations. If the image fails this check, the pipeline either regenerates the image or discards it entirely.

\textbf{Multi-Granularity Evaluation Annotation.}
To support our fine-grained, three-level evaluation framework, the pipeline automatically annotates diagnostic criteria that form the core ground truth for VWG-Bench. For task-level constraints, we define rule dictionaries that delineate the expected behaviors of the foreground (e.g., allowed state changes for manipulated objects), the background (e.g., strict static invariance for unmanipulated areas), and implicit rules (e.g., forbidding mesh interpenetration during object stacking). For sample-level goals, the VLM simulates the outcome of the action to generate linguistic descriptions of the expected final state (i.e., Last-Frame Goal, $\mathcal{G}_{\mathrm{last}}$). For tasks with complex temporal dynamics, intermediate milestones (i.e., Progress Goal, $\mathcal{G}_{\mathrm{prog}}$) are additionally annotated.

\subsection{Evaluation Framework and Protocols}
\label{subsec:evaluation_framework}

Our evaluation framework is designed to systematically assess the reasoning capabilities of video generation models, prioritizing logical coherence and physical adherence over mere visual fidelity. To mitigate evaluation hallucinations, we employ a VLM-as-Judge paradigm, where the judge strictly cross-references generated frames against structured annotations. All metrics across our three-level framework are scored on a 1--5 integer scale (according to the degree of consistency with the corresponding annotations).

\textbf{Video-Level Assessment (Coarse-Grained Quality Control).} 
Before probing reasoning, we must ensure the generated video is a valid visual medium. This foundational filter evaluates Video Quality by assessing holistic visual fidelity and temporal smoothness. The VLM judge explicitly penalizes technical flaws that obscure the video content---such as severe camera jitter, temporal incoherence, unnatural flickering, or excessive static frames---ensuring that poor baseline rendering does not confound the subsequent reasoning evaluation.

\textbf{Task-Level Assessment (Fine-Grained Consistency \& Constraints).} 
This level leverages rule dictionaries to verify whether the model respects spatial boundaries and logical constraints throughout the generated video.
Specifically, Progress Consistency scrutinizes the temporal coherence of dynamic elements by checking whether the foreground evolves logically while penalizing unprompted changes or morphological collapse in the background. Implicit Rule Following assesses adherence to physical laws and hidden boundaries inherent to the task, strictly penalizing violations such as objects interpenetrating solid meshes, reversing gravity, or rigid bodies undergoing unreasonable soft deformations.

\textbf{Sample-Level Assessment (Goal Realization).} 
The final level evaluates the semantic alignment between the temporal states of the video and the user's explicit instructions. Progress Goal evaluates necessary intermediate milestones; for complex temporal progressions, the judge inspects intermediate keyframes to ensure the ``Start $\rightarrow$ Intermediate $\rightarrow$ End'' sequence is correctly chained without skipping or reversing logical steps. Last-Frame Goal assesses ultimate task completion by comparing the terminal state of the generated video against the annotated goal description to determine whether the objective was achieved.

%% file: sections/3_vid_pre.tex
\section{\texttt{Vid-PRE}: A Universal Prompt Reasoner and Enhancer}
\label{sec:vid_pre}

\begin{figure*}[!t]
    \centering
    \includegraphics[width=0.99\linewidth]{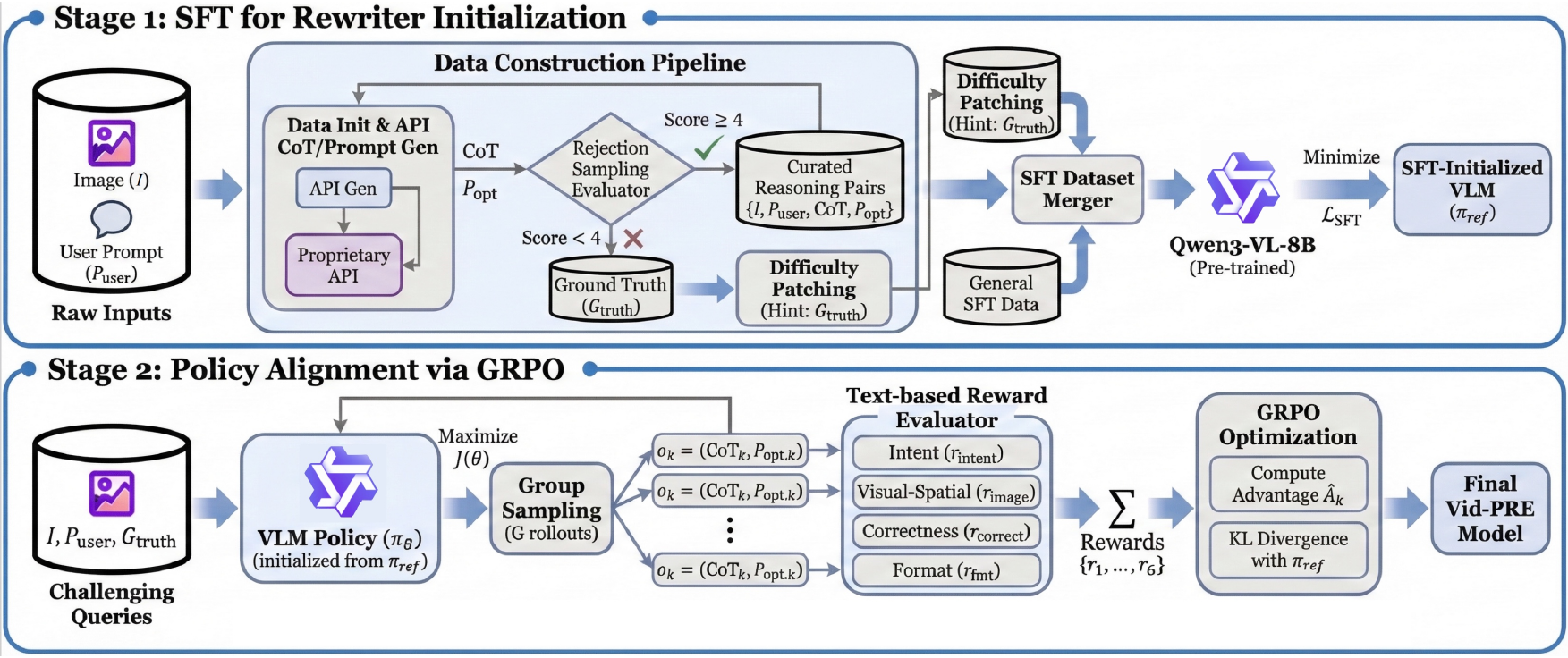}
    \caption{An overview of the training framework for \texttt{Vid-PRE}.}
    \label{fig:method}
\end{figure*}

Current Image-to-Video (I2V) generation models excel at visual rendering but struggle with complex logical derivation and constraint satisfaction. To address this, we introduce Vid-PRE, a model-agnostic prompt rewriting module. The core philosophy of Vid-PRE is a strict division of labor: it offloads the cognitive burden of spatio-temporal reasoning to a dedicated VLM, allowing the downstream video generator to focus exclusively on rendering based on highly explicit and executable instructions.

\subsection{Problem Formulation}
\label{subsec:formulation}

We formulate video prompt rewriting as a conditional text generation problem.
Let $\mathcal{I}$ denote the first-frame image and $\mathcal{P}_{\mathrm{user}}$ the user instruction.
A naive formulation would learn a direct mapping
$(\mathcal{I}, \mathcal{P}_{\mathrm{user}}) \rightarrow \mathcal{P}_{\mathrm{opt}}$,
but this often leads to prompts that lack explicit logical grounding.
To encourage structured reasoning during rewriting, we introduce an intermediate
representation CoT that captures the reasoning process linking
the visual context and the final executable prompt.
The objective of Vid-PRE is therefore to model the joint distribution
$P(\mathrm{CoT}, \mathcal{P}_{\mathrm{opt}} \mid \mathcal{I}, \mathcal{P}_{\mathrm{user}})$.
Let $x=(\mathcal{I},\mathcal{P}_{\mathrm{user}})$ denote the input and
$y=[\mathrm{CoT};\mathcal{P}_{\mathrm{opt}}]$ the concatenated output sequence.
The generation process follows a standard autoregressive factorization:
\setlength{\abovedisplayshortskip}{0pt}
\setlength{\belowdisplayshortskip}{0pt}
\[
P(y \mid x)
=
\prod_{t=1}^{|y|}
P(y_t \mid y_{<t}, x).
\]

This formulation encourages the model to first articulate the reasoning
required to interpret the visual scene and constraints, and then produce
a concise executable prompt $\mathcal{P}_{\mathrm{opt}}$.

\subsection{Stage 1: Supervised Fine-Tuning with Reasoning Data}
\label{subsec:sft}

The first stage initializes the VLM with the structural capability to follow I2V formatting conventions and execute rigorous reasoning. We construct a high-quality dataset of approximately 20K tuples $\{\mathcal{I}, \mathcal{P}_{\mathrm{user}}, \mathrm{CoT}, \mathcal{P}_{\mathrm{opt}}\}$ through an automated rejection sampling paradigm using advanced proprietary APIs. As illustrated in the upper part of Figure~\ref{fig:method}, the data construction process follows four steps:

\textbf{Step1: Data Initialization.}
For each given pair $(\mathcal{I}, \mathcal{P}_{\mathrm{user}})$, we computationally define the rigorous ground truth ($\mathcal{G}_{\mathrm{truth}}$, \eg, the expected last-frame state or progress goal).

\textbf{Step2: CoT and Prompt Generation.}
The API analyzes the initial image, clarifies the user's intent, and identifies
critical constraints to produce a natural-language $\mathrm{CoT}$. It then
generates a concise prompt $\mathcal{P}_{\mathrm{opt}}$ suitable for video
generation.

\textbf{Step3: Rejection Sampling.}
To ensure the reliability of generated samples, we introduce an independent evaluator to assess the quality of each candidate $\mathcal{P}_{\mathrm{opt}}$. The evaluator has access to the input pair $(\mathcal{I}, \mathcal{P}_{\mathrm{user}})$ as well as the corresponding ground truth $\mathcal{G}_{\mathrm{truth}}$, and evaluates the output from two perspectives. First, it verifies visual consistency, checking whether the prompt introduces hallucinated objects, incorrect spatial relations, or directional misunderstandings with respect to the image. Second, it assesses whether the optimized prompt logically leads to the target state defined by $\mathcal{G}_{\mathrm{truth}}$. The evaluator then assigns a comprehensive score on a 1--5 scale. Only samples with scores $\ge 4$ are retained together with their corresponding $\mathrm{CoT}$. This rejection sampling strategy ensures that the reasoning process and the resulting prompt remain logically coherent and consistent.

\textbf{Step4: Difficulty Patching.}
For exceptionally hard task categories where the API fails to deduce a valid logical chain under zero-shot conditions, we relax the constraint by providing $\mathcal{G}_{\mathrm{truth}}$ as a hint, allowing it to generate critical logical patches for the $\mathrm{CoT}$. The patched outputs are then re-evaluated through the same rejection sampling pipeline in Step~(3) to maintain quality control.

These curated reasoning pairs are mixed with general SFT data---to preserve the base conversational capabilities of the VLM---and used to fine-tune a pre-trained Qwen3-VL-8B~\cite{bai2025qwen3} model. The optimization minimizes the standard negative log-likelihood loss:
\setlength{\abovedisplayshortskip}{0pt}
\setlength{\belowdisplayshortskip}{0pt}
\begin{equation}
\mathcal{L}_{\mathrm{SFT}}(\theta)
=
-
\mathbb{E}_{\mathcal{D}_{\mathrm{SFT}}}
\left[
\log
P_\theta
(\mathrm{CoT}, \mathcal{P}_{\mathrm{opt}}
\mid
\mathcal{I}, \mathcal{P}_{\mathrm{user}})
\right].
\end{equation}

\subsection{Stage 2: Policy Alignment via GRPO}
\label{subsec:grpo}

While SFT provides a strong logical initialization, performance on complex, long-horizon reasoning tasks remains suboptimal. As illustrated in the bottom part of Figure~\ref{fig:method}, to further unlock reasoning potential, we apply GRPO using 1.5K specialized prompts collected across 10 challenging tasks that exhibit the highest error rates after the SFT stage.

A key design decision is our adoption of a purely text-based reward system in lieu of video-generation-based rewards. This choice is motivated by three considerations: (1)~\textbf{Stability and Efficiency}: Current video models exhibit substantial aesthetic variance, where poor rendering can easily suppress the reward signal of a logically correct prompt. Text-based evaluation avoids this instability and the considerable computational cost of rendering videos during RL training. (2)~\textbf{Generator Agnosticism}: By decoupling from any specific video generator, Vid-PRE acquires no model-specific priors, functioning as a universal plug-and-play module. (3)~\textbf{Task Nature}: I2V reasoning tasks are tightly bounded by the initial frame and deterministic physical/geometric constraints; thus, optimizing the textual prompt's semantic correctness directly addresses the root cause of reasoning failures.

During training, for each query $x = (\mathcal{I}, \mathcal{P}_{\mathrm{user}}, \mathcal{G}_{\mathrm{truth}})$, the policy $\pi_\theta$ samples a group of $G$ outputs $\{o_1, \dots, o_G\}$, where $o_k = (\mathrm{CoT}_k, \mathcal{P}_{\mathrm{opt},k})$. Our reward function evaluates each rollout along three dimensions on a 1--5 scale. \textbf{Intent Preservation} ($r_{\mathrm{intent}}$) verifies that $\mathcal{P}_{\mathrm{opt},k}$ reasonably extends and refines $\mathcal{P}_{\mathrm{user}}$ without altering the core objective. It ensures key objects and goals are preserved, while minor wording differences (\eg, camera terminology) are not overly penalized. \textbf{Visual-Spatial Consistency} ($r_{\mathrm{image}}$) checks whether the spatial relations described in $\mathcal{P}_{\mathrm{opt},k}$ (\eg, ``bottom right'') accurately reflect the actual content of the image $\mathcal{I}$, strictly penalizing hallucinated visual elements or invalid spatial relationships. \textbf{Solution Correctness} ($r_{\mathrm{correct}}$) evaluates whether the explicit steps inferred in the prompt correctly solve the task and reach $\mathcal{G}_{\mathrm{truth}}$, emphasizing critical problem-solving details (\eg, specific paths or action steps) over generic generation constraints. In addition, a format compliance score ($r_{\mathrm{format}}$) ensures outputs adhere to the prescribed structural template. The final reward is a weighted combination: $r_k = w_1 r_{\mathrm{intent}} + w_2 r_{\mathrm{image}} + w_3 r_{\mathrm{correct}} + w_4 r_{\mathrm{format}}$.

GRPO optimizes the policy by maximizing the advantage of outputs with higher relative rewards within the group~\cite{fang2025inflvg, xue2025dancegrpo, feng2025video}. The advantage for each rollout is computed as the group-normalized reward: $\hat{A}_k = (r_k - \mu_G) / \sigma_G$, where $\mu_G$ and $\sigma_G$ are the mean and standard deviation of rewards within the group $\{r_1, \dots, r_G\}$. Let $\pi_{\mathrm{ref}}$ denote the SFT-initialized policy serving as the fixed reference for KL regularization:
\setlength{\abovedisplayshortskip}{0pt}
\setlength{\belowdisplayshortskip}{0pt}
\begin{equation}
\mathcal{J}(\theta) =
\mathbb{E}_{x \sim \mathcal{D}_{\mathrm{RL}},\; o \sim \pi_{\theta_{\mathrm{old}}}}
\left[
\frac{1}{G}
\sum_{k=1}^{G}
\frac{\pi_\theta(o_k \mid x)}{\pi_{\theta_{\mathrm{old}}}(o_k \mid x)}
\hat{A}_k
-
\beta \, \mathbb{D}_{\mathrm{KL}}(\pi_\theta \| \pi_{\mathrm{ref}})
\right].
\end{equation}

Through this paradigm, Vid-PRE learns to produce logically precise and physically grounded reasoning chains that translate into robust instructions for downstream video generation.

%% file: sections/4_exp.tex
\section{Experiments}
\label{sec:experiments}

\subsection{Settings}
\label{subsec:settings}

We evaluate the general reasoning capability of six video generation models on VWG-Bench, including one open-weight model (Wan2.2) and five commercial systems (Wan2.5, Wan2.6-Flash, Kling-2.5-Turbo-Pro, Sora2, and Veo3.1). For each $(\mathcal{I}, \mathcal{P}_{\mathrm{user}})$ pair, we sample three generations and report the averaged score for each evaluation dimension. During evaluation, we extract the last frame of each generated video to assess $\mathcal{G}_{\mathrm{last}}$, while the remaining metrics are evaluated using a frame sampling configuration of 2 fps with a maximum of 16 frames.

We initialize Vid-PRE with the pre-trained Qwen3-VL-8B~\cite{bai2025qwen3} model, and use Gemini-3.0-Pro as the data generator and the unified evaluator. During the SFT stage, the model is trained on approximately 20K high-quality samples for 3 epochs with a batch size of 128. The maximum image token length is constrained to 1{,}024 to balance fine-grained visual detail and computational efficiency. In the subsequent GRPO stage, the policy is further optimized using 1.5K high-difficulty samples for approximately 3K optimization steps under the text-based reward system described earlier. All training is conducted on 8 H100 GPUs using the DeepSpeed ZeRO-3 optimization framework.

During evaluation, we apply our prompt rewriting method only to models that are either open-weight (Wan2.2) or allow disabling their built-in Prompt Extension module (Wan2.5 and Wan2.6-Flash). For the latter two models, we additionally report results with their native Prompt Extension (PE) enabled to provide a direct comparison. 
Beyond VWG-Bench, we further evaluate Vid-PRE on two additional high-quality open benchmarks. The first is MME-CoF, which contains 59 high-quality $(\text{Image}, \text{Prompt})$ pairs and evaluates generated videos from five aspects: instruction alignment, temporal consistency, visual stability, content fidelity, and focus relevance. 
The second benchmark is V-ReasonBench, which consists of 12 task categories with a total of 327 $(\text{Image}, \text{Prompt}, \text{Ground Truth})$ triplets. Each category is designed to be last-frame verifiable, meaning that task success can be determined solely from the final frame (e.g., Sudoku or arithmetic reasoning tasks). The benchmark groups the 12 categories into four reasoning dimensions: Structured Problem-Solving, Spatial Cognition, Pattern-based Inference, and Physical Dynamics. We report the mean score over five random seeds for each dimension.
\subsection{Main Results}
\label{subsec:main_results}

\textbf{Benchmarking Current Models on \texttt{VWG-Bench}.}
We first establish a baseline diagnostic using VWG-Bench (Table~\ref{tab:vwg_baseline}). Even the strongest commercial model (Veo3.1) achieves only 3.55 overall on a 1--5 scale, underscoring the challenge of our benchmark. A consistent pattern emerges across all models: video quality scores are substantially higher than reasoning-intensive metrics. For example, Wan2.2 attains 2.32 in video quality but only 1.56 in implicit rule following, and even Sora2 drops from 3.45 to 2.98 between video quality and last-frame goal accuracy. This confirms our central hypothesis: current models excel at visual rendering but struggle to maintain physical constraints, logical rules, and temporal causality without explicit textual guidance.

\textbf{Enhancing Reasoning with \texttt{Vid-PRE}.}
We evaluate Vid-PRE on two external benchmarks---MME-CoF and V-\allowbreak{}ReasonBench---as well as our VWG-Bench (Table~\ref{tab:external_benchmarks}). Integrating Vid-PRE yields substantial improvements across all models without any architectural modifications to the video generators.
On MME-CoF (Table~\ref{tab:external_mme}), Vid-PRE boosts Wan2.2's overall score from 1.72 to 2.58, a 50\% gain, with instruction alignment and content fidelity showing the largest absolute gains. It also outperforms Wan~PE, raising Wan2.5 from 2.84 to 3.07 and Wan2.6-\allowbreak{}Flash from 2.82 to 3.09.
On V-\allowbreak{}ReasonBench, the advantage is even more pronounced
(Table~\ref{tab:external_vreason}), whose tasks demand fine-grained physical
and spatial reasoning. Vid-PRE improves Wan2.2's average accuracy from 26.65
to 44.48, a 67\% gain, with spatial cognition increasing from 1.32 to 36.25.
Against Wan~PE, Vid-PRE delivers an additional $+25\%$ gain on Wan2.5 and
$+33\%$ on Wan2.6-Flash, confirming that generic prompt rewriting is no
substitute for structured, constraint-aware reasoning.
On VWG-Bench (Table~\ref{tab:external_vwg}), Vid-PRE raises Wan2.2 from 2.13 to 2.51, an 18\% gain, and improves its weakest baseline dimension, implicit rule following, from 1.56 to 2.12. For Wan2.6-Flash, Vid-PRE achieves 3.26, surpassing the Wan~PE baseline of 3.04 and narrowing the gap to commercial systems like Sora2 (3.46) and Veo3.1 (3.55).

% --------------------------------------------------------
% Table 1: VWG-Bench Baseline Diagnostic
% --------------------------------------------------------
\begin{table}[t]
\centering
\caption{Baseline diagnostic on \texttt{VWG-Bench}. All models are evaluated without prompt extension. Commercial models are included to contextualize task difficulty.}
\label{tab:vwg_baseline}
\resizebox{\columnwidth}{!}{
\begin{tabular}{lcccccc}
\toprule
\textbf{Model} & \textbf{Video Quality} & \textbf{Prog.~Consist.} & \textbf{Rule Follow.} & \textbf{Prog.~Goal} & \textbf{Last Goal} & \textbf{Overall} \\
\midrule
Wan2.2              & 2.32 & 2.33 & 1.56 & 2.45 & 1.98 & 2.13 \\
Wan2.5             & 2.70 & 2.86 & 3.14 & 2.48 & 2.56 & 2.75 \\
Wan2.6-Flash        & 2.73 & 2.92 & 3.23 & 2.50 & 2.62 & 2.80 \\
Kling-2.5-Turbo-Pro & 2.95 & 3.12 & 3.38 & 2.62 & 2.78 & 2.97 \\
Sora2               & 3.45 & 3.68 & 3.79 & 3.32 & 2.98 & 3.44 \\
Veo3.1              & \textbf{3.54} & \textbf{3.78} & \textbf{3.95} & \textbf{3.45} & \textbf{3.05} & \textbf{3.55} \\
\bottomrule
\end{tabular}
}
\end{table}

\vspace{-4mm}

% --------------------------------------------------------
% Table 2: Generalization across Benchmarks
% --------------------------------------------------------
\begin{table*}[t]
\centering
\caption{Performance of \texttt{Vid-PRE} across reasoning benchmarks. Wan~PE denotes the model's built-in prompt extension. \texttt{Vid-PRE} consistently outperforms both the base models and their native PE across all metrics.}
\label{tab:external_benchmarks}

\vspace{-4mm}

% Sub-table A: MME-CoF
\begin{subtable}[t]{\textwidth}
\centering
\caption{MME-CoF.}
\label{tab:external_mme}
\resizebox{\textwidth}{!}{
\begin{tabular}{llcccccc}
\toprule
\textbf{Model} & \textbf{Prompt Extension} & \textbf{Instr.~Align.} & \textbf{Temp.~Consist.} & \textbf{Vis.~Stability} & \textbf{Content Fidelity} & \textbf{Focus Relevance} & \textbf{Overall Score} \\
\midrule
\multirow{2}{*}{Wan2.2}
  & None              & 0.69 & 2.36 & 2.63 & 1.16 & 1.77 & 1.72 \\
  & \cellcolor{gray!10}\texttt{Vid-PRE} & \cellcolor{gray!10}\textbf{1.65} & \cellcolor{gray!10}\textbf{2.95} & \cellcolor{gray!10}\textbf{3.11} & \cellcolor{gray!10}\textbf{2.38} & \cellcolor{gray!10}\textbf{2.82} & \cellcolor{gray!10}\textbf{2.58} \\
\midrule
\multirow{3}{*}{Wan2.5}
  & None              & 1.57 & 2.85 & 3.34 & 2.45 & 2.87 & 2.62 \\
  & Wan PE            & 1.98 & 2.88 & 3.41 & 2.59 & 3.35 & 2.84 \\
  & \cellcolor{gray!10}\texttt{Vid-PRE} & \cellcolor{gray!10}\textbf{2.34} & \cellcolor{gray!10}\textbf{3.28} & \cellcolor{gray!10}\textbf{3.59} & \cellcolor{gray!10}\textbf{2.69} & \cellcolor{gray!10}\textbf{3.47} & \cellcolor{gray!10}\textbf{3.07} \\
\midrule
\multirow{3}{*}{Wan2.6-Flash}
  & None              & 1.67 & 3.05 & 3.39 & 2.49 & 2.85 & 2.69 \\
  & Wan PE            & 2.00 & 2.85 & 3.34 & 2.61 & 3.31 & 2.82 \\
  & \cellcolor{gray!10}\texttt{Vid-PRE} & \cellcolor{gray!10}\textbf{2.25} & \cellcolor{gray!10}\textbf{3.26} & \cellcolor{gray!10}\textbf{3.62} & \cellcolor{gray!10}\textbf{2.86} & \cellcolor{gray!10}\textbf{3.43} & \cellcolor{gray!10}\textbf{3.09} \\
\bottomrule
\end{tabular}
}
\end{subtable}

% Sub-table B: V-ReasonBench
\begin{subtable}[t]{\textwidth}
\centering
\caption{V-ReasonBench.}
\label{tab:external_vreason}
\resizebox{\textwidth}{!}{
\begin{tabular}{llccccc}
\toprule
\textbf{Model} & \textbf{Prompt Extension} & \textbf{Struct.~Solving} & \textbf{Spatial Cognition} & \textbf{Pattern Inference} & \textbf{Physical Dynamics} & \textbf{Average} \\
\midrule
\multirow{2}{*}{Wan2.2}
  & None              & 29.27 & 1.32  & 45.12 & 30.90 & 26.65 \\
  & \cellcolor{gray!10}\texttt{Vid-PRE} & \cellcolor{gray!10}\textbf{33.90} & \cellcolor{gray!10}\textbf{36.25} & \cellcolor{gray!10}\textbf{60.49} & \cellcolor{gray!10}\textbf{47.29} & \cellcolor{gray!10}\textbf{44.48} \\
\midrule
\multirow{3}{*}{Wan2.5}
  & None              & 25.95 & 20.05 & 46.30 & 33.05 & 31.34 \\
  & Wan PE            & 35.55 & 38.25 & 52.30 & 38.15 & 41.06 \\
  & \cellcolor{gray!10}\texttt{Vid-PRE} & \cellcolor{gray!10}\textbf{44.05} & \cellcolor{gray!10}\textbf{42.35} & \cellcolor{gray!10}\textbf{66.15} & \cellcolor{gray!10}\textbf{52.20} & \cellcolor{gray!10}\textbf{51.19} \\
\midrule
\multirow{3}{*}{Wan2.6-Flash}
  & None              & 26.85 & 20.71 & 47.10 & 33.74 & 32.10 \\
  & Wan PE            & 36.57 & 29.18 & 53.23 & 39.02 & 39.50 \\
  & \cellcolor{gray!10}\texttt{Vid-PRE} & \cellcolor{gray!10}\textbf{45.46} & \cellcolor{gray!10}\textbf{43.78} & \cellcolor{gray!10}\textbf{67.35} & \cellcolor{gray!10}\textbf{53.68} & \cellcolor{gray!10}\textbf{52.57} \\
\bottomrule
\end{tabular}
}
\end{subtable}

% Sub-table C: VWG-Bench
\begin{subtable}[t]{\textwidth}
\centering
\caption{VWG-Bench.}
\label{tab:external_vwg}
\resizebox{\textwidth}{!}{
\begin{tabular}{llcccccc}
\toprule
\textbf{Model} & \textbf{Prompt Extension} & \textbf{Video Quality} & \textbf{Prog.~Consist.} & \textbf{Rule Follow.} & \textbf{Prog.~Goal} & \textbf{Last Goal} & \textbf{Overall} \\
\midrule
\multirow{2}{*}{Wan2.2}
  & None              & 2.32 & 2.33 & 1.56 & 2.45 & 1.98 & 2.13 \\
  & \cellcolor{gray!10}\texttt{Vid-PRE} & \cellcolor{gray!10}\textbf{2.65} & \cellcolor{gray!10}\textbf{2.68} & \cellcolor{gray!10}\textbf{2.12} & \cellcolor{gray!10}\textbf{2.75} & \cellcolor{gray!10}\textbf{2.35} & \cellcolor{gray!10}\textbf{2.51} \\
\midrule
\multirow{3}{*}{Wan2.5}
  & None              & 2.70 & 2.86 & 3.14 & 2.48 & 2.56 & 2.75 \\
  & Wan PE            & 3.01 & 3.18 & 3.35 & 2.66 & 2.74 & 2.99 \\
  & \cellcolor{gray!10}\texttt{Vid-PRE} & \cellcolor{gray!10}\textbf{3.25} & \cellcolor{gray!10}\textbf{3.41} & \cellcolor{gray!10}\textbf{3.55} & \cellcolor{gray!10}\textbf{2.83} & \cellcolor{gray!10}\textbf{2.93} & \cellcolor{gray!10}\textbf{3.19} \\
\midrule
\multirow{3}{*}{Wan2.6-Flash}
  & None              & 2.73 & 2.92 & 3.23 & 2.50 & 2.62 & 2.80 \\
  & Wan PE            & 3.05 & 3.18 & 3.45 & 2.68 & 2.86 & 3.04 \\
  & \cellcolor{gray!10}\texttt{Vid-PRE} & \cellcolor{gray!10}\textbf{3.28} & \cellcolor{gray!10}\textbf{3.49} & \cellcolor{gray!10}\textbf{3.62} & \cellcolor{gray!10}\textbf{2.88} & \cellcolor{gray!10}\textbf{3.02} & \cellcolor{gray!10}\textbf{3.26} \\
\bottomrule
\end{tabular}
}
\end{subtable}

\end{table*}

\subsection{Analysis and Discussion}
\label{subsec:analysis}

\textbf{Evaluator Reliability.}
We validate our VLM-as-Judge protocol by comparing automatic scores with human judgments. Specifically, we sample 100 prompts across all nine VWG-Bench dimensions and evaluate 200 videos generated by Veo3.1 and Wan2.6-Flash. Three human experts independently score each video on the same 1--5 criteria used by the automatic evaluator. As shown in Table~\ref{tab:judge_reliability}, Spearman's $\rho$ ranges from 0.69 to 0.80 and the mean absolute error remains below 0.55 across all criteria. This agreement indicates that the automatic scores track human-perceived reasoning quality rather than only superficial rendering preference, supporting the reliability of the benchmark-level model comparisons.

\begin{table}[htbp]
\centering
\caption{Agreement between human experts and Gemini-3.0-Pro on the VWG-Bench evaluation criteria.}
\label{tab:judge_reliability}
\small
\setlength{\tabcolsep}{5pt}
\begin{tabular}{lcccc}
\toprule
\textbf{Criterion} & \textbf{Human} & \textbf{Gemini} &
\textbf{$\rho$\,$\uparrow$} & \textbf{MAE\,$\downarrow$} \\
\midrule
Video Quality        & 3.20 & 3.44 & 0.69 & 0.54 \\
Progress Consistency & 3.42 & 3.35 & 0.78 & 0.42 \\
Rule Following       & 3.47 & 3.59 & 0.73 & 0.48 \\
Progress Goal        & 3.06 & 2.98 & 0.75 & 0.41 \\
Last-Frame Goal      & 2.93 & 2.84 & 0.80 & 0.39 \\
\bottomrule
\end{tabular}
\end{table}

\textbf{Ablation Study.}
We ablate the contribution of each training stage on both external benchmarks using Wan2.2 and Wan2.6-Flash as base generators (Table~\ref{tab:ablation_sft_rl}).
SFT alone already provides a significant lift over the baselines, showing that structured reasoning and constraint-aware prompt generation are effective even without reinforcement learning. The subsequent GRPO stage consistently pushes performance further, as our text-based reward system explicitly penalizes spatial hallucinations and logical misalignments. The same trend holds for Wan2.6-Flash. These results confirm that the two stages are complementary: SFT provides the structural foundation, and GRPO refines the policy toward tighter constraint satisfaction.
\begin{table}[htbp]
\centering
\caption{Ablation on training stages.}
\label{tab:ablation_sft_rl}
\scriptsize
\setlength{\tabcolsep}{4pt}
\begin{tabular}{llcc}
\toprule
\textbf{Model} & \textbf{Variant} & \textbf{MME-CoF} & \textbf{V-ReasonBench} \\
\midrule
\multirow{3}{*}{Wan2.2}
  & Base                                    & 1.72 & 26.65 \\
  & + \texttt{Vid-PRE}\textsubscript{SFT}   & 2.44 & 40.26 \\
  & \cellcolor{gray!10}+ \texttt{Vid-PRE}\textsubscript{SFT+RL} & \cellcolor{gray!10}\textbf{2.58} & \cellcolor{gray!10}\textbf{44.48} \\
\midrule
\multirow{3}{*}{Wan2.6-Flash}
  & Base                                    & 2.69 & 32.10 \\
  & + \texttt{Vid-PRE}\textsubscript{SFT}   & 2.97 & 45.96 \\
  & \cellcolor{gray!10}+ \texttt{Vid-PRE}\textsubscript{SFT+RL} & \cellcolor{gray!10}\textbf{3.09} & \cellcolor{gray!10}\textbf{52.57} \\
\bottomrule
\end{tabular}
\end{table}

\begin{figure*}[!t]
    \centering
\includegraphics[width=0.99\linewidth]{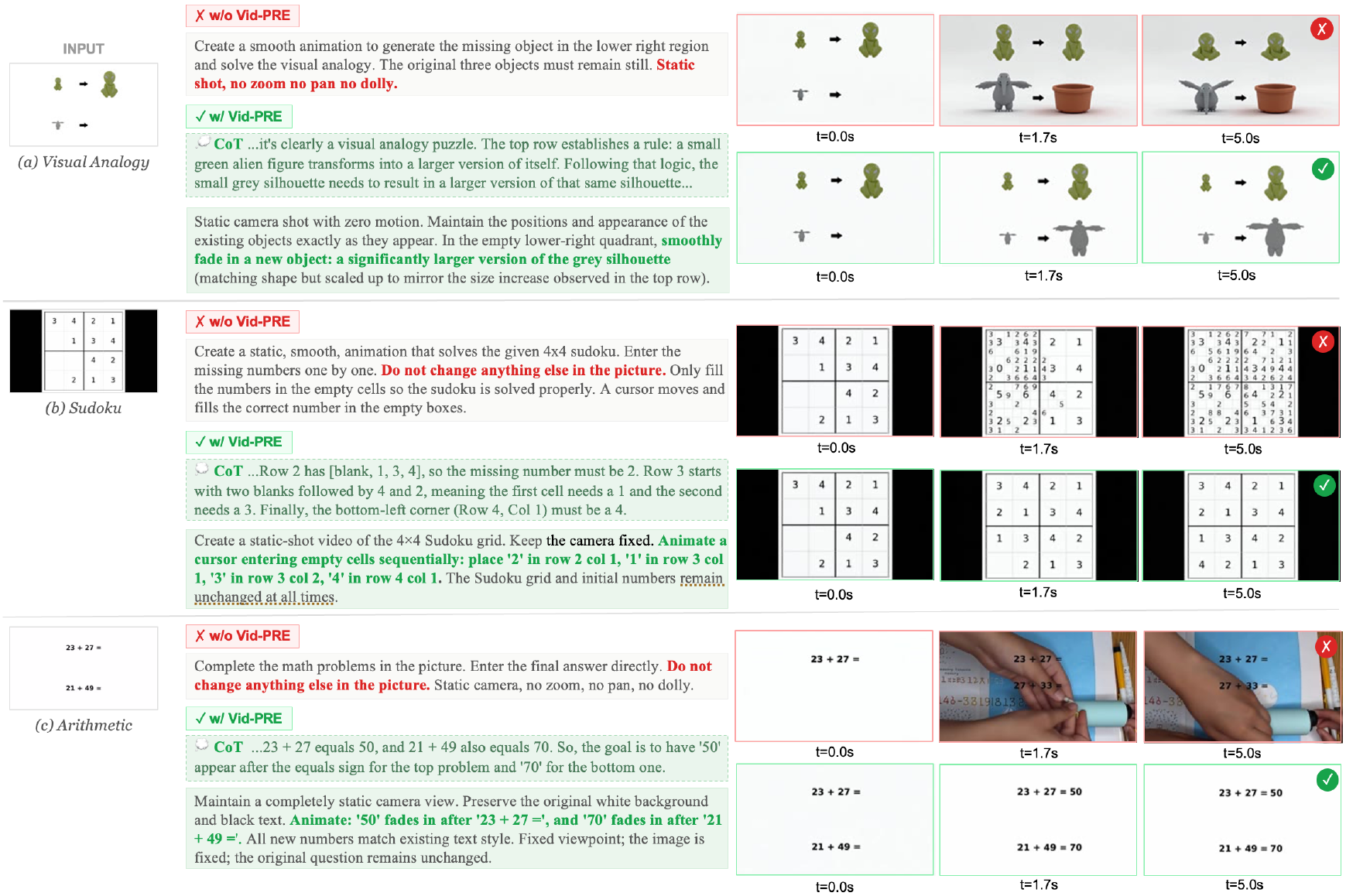}
    \caption{Qualitative comparison of video generation with and without \texttt{Vid-PRE}. While generic constraints (red) fail to prevent logical errors and visual hallucinations in base models, \texttt{Vid-PRE} translates explicit reasoning into precise visual instructions (green), significantly improving both reasoning accuracy and video quality.}
    \label{fig:qualitative}
\end{figure*}

\textbf{Qualitative Analysis.}
Figure~\ref{fig:qualitative} presents a qualitative comparison of video generation with and without Vid-PRE across diverse reasoning tasks. As illustrated, base video models fundamentally lack the capacity for zero-shot logical deduction. When given standard prompts, even those explicitly containing strict negative constraints, the base models suffer from severe reasoning failures and visual degradation. For instance, they hallucinate unrelated objects in visual analogies.
By contrast, Vid-PRE dramatically enhances both reasoning capability and generation quality by transforming abstract logical problems into explicit, renderable visual instructions. It first employs a CoT to autonomously deduce the correct logical solutions and then rewrites the prompt to provide precise, step-by-step visual guidance. Consequently, Vid-PRE ensures strict physical constraint compliance, prevents visual hallucinations, and successfully executes complex think-with-video tasks.

%% file: sections/5_related_work.tex
\section{Related Work}
\label{sec:related_work}

\textbf{Image-to-Video Generation.}
I2V generation has progressed rapidly, driven primarily by diffusion and transformer architectures. Since Stable Video Diffusion~\cite{blattmann2023stable} established the paradigm of image-conditioned generation, subsequent works have scaled this approach considerably. Open-source frameworks such as CogVideoX~\cite{yang2024cogvideox}, Wan~\cite{wan2025wan}, and HunyuanVideo~\cite{kong2024hunyuanvideo} have democratized high-resolution synthesis through advanced expert transformers and 3D causal VAEs, while commercial systems like Sora~\cite{openai2025sora2}, Veo~\cite{GoogleDeepMind2025Veo3}, Kling~\cite{kuaishou2024kling}, and Seedance~\cite{gao2025seedance} have achieved photorealistic fidelity and remarkable temporal consistency. Despite these advances in rendering visual dynamics, their capacity for structured, rule-bound reasoning remains fundamentally underexplored.

\textbf{Prompt Rewriting and Optimization.}
The distributional gap between concise user inputs and the detailed captions used during model training makes prompt rewriting critical for video generation. Early approaches bridge this gap via zero-shot in-context learning using LLMs (\eg, GLM-4 in CogVideoX~\cite{yang2024cogvideox} or GPT-4o in Open-Sora~\cite{opensora}). To mitigate hallucination and intent drift inherent in zero-shot rewriting, recent methods have shifted toward reward-guided optimization, employing techniques such as Direct Preference Optimization (Prompt-A-Video~\cite{ji2025prompt}), multi-level feedback (VPO~\cite{cheng2025vpo}), and retrieval-augmented refinement (RAPO~\cite{gao2025devil}). However, these techniques predominantly focus on enhancing aesthetic quality and generic text-video alignment. In contrast, Vid-PRE specifically targets reasoning enhancement by offloading the cognitive burden to a VLM and training with purely text-based reinforcement learning signals.

\textbf{Benchmarks for I2V Models.}
Existing evaluation frameworks broadly fall into two categories.
The first focuses on perceptual quality and general generation capabilities:
benchmarks like VBench/VBench++~\cite{huang2024vbench,huang2025vbench++} and
WorldScore~\cite{duan2025worldscore} assess visual fidelity and scene dynamics
but do not probe higher-order logical reasoning.
The second category, motivated by recent demonstrations of reasoning in video
generation~\cite{wiedemer2025video}, evaluates the cognitive limits of video
models.
Several broad benchmarks have emerged (\eg, MME-CoF~\cite{guo2025video},
VideoThinkBench~\cite{tong2025thinking}, TiViBench~\cite{chen2025tivibench}),
alongside specialized ones targeting procedural pathfinding
(VR-Bench~\cite{yang2025reasoning}), strict rule execution
(RULER-Bench~\cite{he2025ruler}), and implicit physical and visual rules
(\eg, V-ReasonBench~\cite{luo2025v}, RISE-Video~\cite{liu2026rise},
MMGR~\cite{cai2025mmgr}), as shown in
Table~\ref{tab:benchmark_comparison}.
Despite advancing the evaluation landscape, these benchmarks often isolate
specific cognitive skills or rely on holistic, coarse-grained scoring.
VWG-Bench addresses these limitations with a comprehensive taxonomy spanning
9 dimensions and 38 tasks, equipped with a fine-grained three-level
VLM-as-Judge framework.

%% file: sections/6_conclusion_future_work.tex
\section{Conclusion}
\label{sec:conclusion}
In this work, we introduced \texttt{VWG-Bench}, a comprehensive evaluation
framework designed to rigorously assess the think-with-video reasoning
capabilities of modern video generative models.
Through our three-level VLM-as-Judge evaluation, we revealed a persistent gap:
current models excel at visual rendering but struggle with logic-heavy,
rule-constrained, and long-horizon tasks.
To bridge this gap, we proposed \texttt{Vid-PRE}, a model-agnostic prompt
enhancer that offloads spatio-temporal reasoning to a dedicated VLM and
translates raw instructions into constraint-aware, executable prompts.
Trained via SFT and GRPO with purely text-based rewards, \texttt{Vid-PRE}
yields substantial and consistent improvements in reasoning accuracy across
multiple state-of-the-art generators.

% \section{Limitations and Future Work}
% Our current evaluation primarily focuses on open-weight models from the Wan family; validating \texttt{Vid-PRE} on a broader range of architectures (\eg, CogVideoX, HunyuanVideo) would further strengthen the generalizability claims. Additionally, the text-based reward system, while efficient and stable, cannot capture subtle visual artifacts that only manifest in the rendered video. Future work will explore expanding to more complex interactive embodied scenarios, incorporating lightweight video-aware reward signals, and investigating tighter end-to-end integration between reasoning modules and rendering engines.

%% file: sections/supp_additional_evaluation_details.tex
\section{Additional Evaluation Details}

\subsection{Physics Generalization and Per-Dimension Analysis}

Table~\ref{tab:supp_physics} reports the physics-focused results, and
Table~\ref{tab:supp_dimension} provides the complete dimension-level
breakdown.

\begin{table}[H]
\centering
\caption{Vid-PRE on physics-related reasoning tasks.}
\label{tab:supp_physics}
\setlength{\tabcolsep}{5pt}
\resizebox{\textwidth}{!}{%
\begin{tabular}{lcc}
\toprule
\textbf{Model} & \textbf{Physics (MME-CoF)} &
\textbf{Physical Dynamics (V-RB)} \\
\midrule
Wan2.2                    & 2.25 & 30.90 \\
Wan2.2 with Vid-PRE       & \textbf{3.31} ($+47\%$) &
                            \textbf{47.29} ($+53\%$) \\
Wan2.6-Flash              & 3.25 & 33.74 \\
Wan2.6-Flash with Vid-PRE & \textbf{3.45} ($+6\%$) &
                            \textbf{53.68} ($+59\%$) \\
\bottomrule
\end{tabular}
}
\end{table}

\begin{table}[H]
\centering
\caption{Per-dimension breakdown with Vid-PRE on VWG-Bench. Values in
parentheses are relative gains over the corresponding base generator.
``--'' indicates that a dimension has no rule-following annotation.}
\label{tab:supp_dimension}
\setlength{\tabcolsep}{4pt}
\small
\resizebox{\textwidth}{!}{%
\begin{tabular}{lccccc}
\toprule
\textbf{Dimension} & \textbf{VQ} & \textbf{PC} & \textbf{RF} &
\textbf{PG} & \textbf{LG} \\
\midrule
Symbolic       & 2.91 ($+9.7\%$)  & 2.36 ($+13.5\%$) & -- &
                 2.94 ($+17.6\%$) & 2.93 ($+16.7\%$) \\
Graphical      & 3.43 ($+14.3\%$) & 3.35 ($+9.5\%$)  &
                 4.18 ($+11.5\%$) & 4.39 ($+15.5\%$) &
                 3.05 ($+16.4\%$) \\
Embodied       & 3.96 ($+4.2\%$)  & 4.73 ($+2.8\%$)  &
                 4.87 ($+3.6\%$)  & 4.49 ($+5.4\%$)  &
                 3.86 ($+5.5\%$) \\
GUI            & 3.32 ($+14.5\%$) & 3.29 ($+15.4\%$) & -- &
                 4.25 ($+16.4\%$) & 3.02 ($+18.4\%$) \\
Counterfactual & 4.32 ($+3.4\%$)  & 4.84 ($+2.7\%$)  & -- &
                 2.08 ($+6.4\%$)  & 3.54 ($+4.8\%$) \\
Spatial        & 3.86 ($+5.1\%$)  & 4.15 ($+5.2\%$)  &
                 1.78 ($+9.5\%$)  & 3.89 ($+6.1\%$)  &
                 2.72 ($+7.0\%$) \\
Temporal       & 4.06 ($+4.2\%$)  & 4.31 ($+5.0\%$)  & -- &
                 4.32 ($+5.3\%$)  & 4.05 ($+4.9\%$) \\
Physical       & 3.87 ($+5.3\%$)  & 4.58 ($+5.9\%$)  & -- &
                 3.36 ($+5.4\%$)  & 3.06 ($+5.0\%$) \\
Social         & 3.74 ($+3.6\%$)  & 4.32 ($+5.0\%$)  & -- &
                 4.32 ($+6.0\%$)  & 3.83 ($+5.9\%$) \\
\bottomrule
\end{tabular}
}
\end{table}

\subsection{Additional Commercial Model and Reward Sensitivity}

Table~\ref{tab:supp_seedance} compares Seedance~2.0 with Veo3.1.

\begin{table}[H]
\centering
\caption{Seedance~2.0 and Veo3.1 on VWG-Bench.}
\label{tab:supp_seedance}
\setlength{\tabcolsep}{4pt}
\begin{tabular}{lcccccc}
\toprule
\textbf{Model} & \textbf{VQ} & \textbf{PC} & \textbf{RF} &
\textbf{PG} & \textbf{LG} & \textbf{Overall} \\
\midrule
Veo3.1      & 3.54 & \textbf{3.78} & \textbf{3.95} &
              \textbf{3.45} & 3.05 & \textbf{3.55} \\
Seedance2.0 & \textbf{3.67} & 3.52 & 3.87 & 3.38 &
              \textbf{3.17} & 3.52 \\
\bottomrule
\end{tabular}
\end{table}

GRPO uses four reward components; format compliance is always active to
ensure output parsability. Table~\ref{tab:supp_reward} ablates the remaining
three components. Solution correctness is the main source of improvement,
while intent preservation and image consistency provide complementary gains.

\begin{table}[H]
\centering
\caption{Reward-component ablation using Wan2.2.}
\label{tab:supp_reward}
\setlength{\tabcolsep}{5pt}
\begin{tabular}{ccccc}
\toprule
$r_{\mathrm{intent}}$ & $r_{\mathrm{image}}$ &
$r_{\mathrm{correct}}$ & \textbf{MME-CoF} &
\textbf{V-ReasonBench} \\
\midrule
\ding{55} & \ding{55} & \ding{55} & 2.44 & 40.26 \\
\ding{55} & \ding{51} & \ding{55} & 2.40 & 37.85 \\
\ding{51} & \ding{55} & \ding{55} & 2.42 & 39.10 \\
\ding{51} & \ding{51} & \ding{55} & 2.41 & 38.45 \\
\ding{55} & \ding{55} & \ding{51} & 2.53 & 43.35 \\
\ding{51} & \ding{55} & \ding{51} & 2.56 & 44.05 \\
\ding{55} & \ding{51} & \ding{51} & 2.55 & 43.75 \\
\ding{51} & \ding{51} & \ding{51} & \textbf{2.58} & \textbf{44.48} \\
\bottomrule
\end{tabular}
\end{table}

\subsection{Inference Cost}

Vid-PRE prompt rewriting takes approximately 3--5 seconds per sample, whereas
video generation typically takes more than 60 seconds in the evaluated I2V
pipelines. The resulting overhead is below 8\% of end-to-end latency.